\documentclass{article}

\usepackage[preprint,nonatbib]{neurips_2023}

\usepackage[utf8]{inputenc} % allow utf-8 input
\usepackage[T1]{fontenc}    % use 8-bit T1 fonts
\usepackage{hyperref}       % hyperlinks
\usepackage{url}            % simple URL typesetting
\usepackage{booktabs}       % professional-quality tables
\usepackage{amsfonts}       % blackboard math symbols
\usepackage{nicefrac}       % compact symbols for 1/2, etc.
\usepackage{microtype}      % microtypography
\usepackage{xcolor}         % colors
\usepackage{graphicx}
\usepackage{multirow}
\usepackage{caption}
\title{On the Robustness of Latent Diffusion Models}

\author{Jianping Zhang$ ^{1} $ \qquad
Zhuoer Xu$ ^{2} $ \qquad
Shiwen Cui$ ^{2} $ \qquad
Changhua Meng$ ^{2} $ \qquad
\\
Weibin Wu$ ^{3} $ \qquad
Michael R. Lyu$ ^{1} $
\\
$ ^{1} $Department of Computer Science and Engineering, The Chinese University of Hong Kong
\\
$ ^{2} $Tiansuan Lab, Antgroup
\\
$ ^{3} $School of Software Engineering, Sun Yat-sen University
\\
{\tt\small \{jpzhang, lyu\}@cse.cuhk.edu.hk, wuwb36@mail.sysu.edu.cn}
\\
{\tt\small \{xuzhuoer.xze, donn.csw, changhua.mch\}@antgroup.com}
}

\begin{document}

\maketitle

\begin{abstract}
  Latent diffusion models achieve state-of-the-art performance on a variety of generative tasks, such as image synthesis and image editing. However, the robustness of latent diffusion models is not well studied. Previous works only focus on the adversarial attacks against the encoder or the output image under white-box settings, regardless of the denoising process. Therefore, in this paper, we aim to analyze the robustness of latent diffusion models more thoroughly. We first study the influence of the components inside latent diffusion models on their white-box robustness. In addition to white-box scenarios, we evaluate the black-box robustness of latent diffusion models via transfer attacks, where we consider both prompt-transfer and model-transfer settings and possible defense mechanisms. However, all these explorations need a comprehensive benchmark dataset, which is missing in the literature. Therefore, to facilitate the research of the robustness of latent diffusion models, we propose two automatic dataset construction pipelines for two kinds of image editing models and release the whole dataset. Our code and dataset are available at \url{https://github.com/jpzhang1810/LDM-Robustness}.
  \end{abstract}

\section{Introduction}

Diffusion models achieve state-of-the-art performance on a variety of generative tasks, such as image synthesis and image editing \cite{saharia2022photorealistic, ho2020denoising}. Among various diffusion models, latent diffusion models \cite{rombach2022high} stand out for their efficiency in generating or editing high-quality images via an iterative denoising process in the latent space. A latent noise is sampled from a uniform Gaussian distribution and denoised through the diffusion model step by step to form a real image. Furthermore, latent diffusion models enable image editing through conditioning, which adds the embedding of a given image on the sampled latent noise to do image editing. The image editing diffusion models have been widely deployed in real-world applications such as DALL$\cdot$E2\cite{ramesh2022hierarchical} and Stable Diffusion \cite{rombach2022high}. Therefore, we focus on the image editing diffusion models in this paper.

Despite the dazzling performance of latent diffusion models, recent studies show that they are susceptible to adversarial attacks \cite{salman2023raising}, which add human-imperceptible noise to a clean image so that latent diffusion models will incorrectly edit the input image or generate low-quality images. Therefore, for the sake of producing better diffusion models, it is important to analyze the robustness of latent diffusion models.

However, the robustness of the diffusion models has not been well studied in the literature. First of all, previous studies only focus on attacking the encoder \cite{zhuang2023pilot} or distorting the output image \cite{salman2023raising}, without taking the core structure of diffusion models (the denoising process) into consideration. Second, prior works only pay attention to the white-box setting, where attackers can obtain the full information of the victim model, like model structures and weights \cite{goodfellow2014explaining, kurakin2016adversarial, madry2017towards}. However, they neglect the black-box setting, which assumes that attackers do not know the internals of the models and is closer to the real-world settings \cite{dong2018boosting}. Lastly, all these robustness explorations need a high-quality and publicly available benchmark, which is still missing in the literature.

%In general, adversarial attacks can be categorized into two classes: white-box attack \cite{goodfellow2014explaining, kurakin2016adversarial, madry2017towards} and black-box attack \cite{dong2018boosting} based on the knowledge of the victim model. White-box attacker can obtain the full information of the victim model, like model structure and weight, while black-box attacker cannot.

Therefore, we endeavor to solve the aforementioned problems in this paper. To evaluate the robustness of latent diffusion models, we deploy adversarial attacks under various white-box and black-box settings. To facilitate the research on the robustness of diffusion models, we propose two automatic dataset construction pipelines to generate test cases for two categories of image editing diffusion models. The contributions of our work are threefold:

\begin{itemize}
    \item We launch adversarial attacks on representative modules inside the latent diffusion models to analyze their adversarial vulnerability. In addition, we compare the white-box robustness of two categories of latent diffusion models for image editing. 
    
    \item We also study the black-box robustness of latent diffusion models via transfer attacks, where we consider both prompt-transfer and model-transfer settings and possible defense mechanisms.

    \item We propose automatic dataset construction pipelines for analyzing the robustness of two kinds of image editing latent diffusion models. The dataset consists of 500 data pairs (image + prompt) for image variation models and another 500 data triplets (image + prompt + mask) for image inpainting models. We will make the code and the built dataset publicly available.
\end{itemize}

\section{Preliminary}

\subsection{Diffusion Models}

Diffusion models \cite{ho2020denoising, rombach2022high} displaced GANs \cite{goodfellow2020generative, arjovsky2017wasserstein} recently with state-of-the-art performance on generating realistic images. Specifically, diffusion models are capable of generating images or editing images via textual prompts.

The core of the diffusion models is the diffusion process, which is a stochastic differential denoising process. We suppose the image $x$ is from the real image distribution, and the diffusion process gradually adds a small Gaussian noise on the image for $T$ steps.The image of the $t+1$-th step $X_{t+1}$ is exactly $\alpha_t x_t + \beta_t \epsilon_t$, where $\epsilon_t$ follows a Gaussian distribution and $\alpha_t$ as well as $\beta_t$ are the parameters. By gradually adding noise on the image, the image will approximate the Gaussian distribution $\mathcal{N} (0,I)$. The denoising process is to reverse the adding noise process to generate real images starting from the uniform Gaussian distribution. Therefore the network of the diffusion models predicts the noise added on the step $t$ with the input $x_{t+1}$ to remove the noise.

\begin{figure}[t]
  \centering
  \centerline{\includegraphics[width=0.8\linewidth]{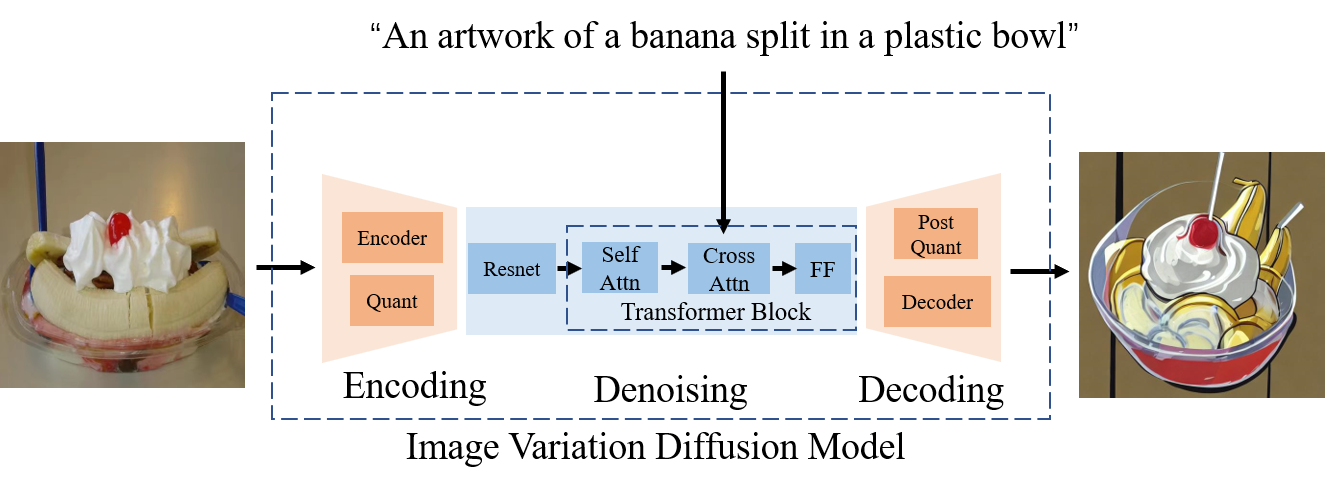}}
  \caption{
  The workflow of image variation latent diffusion model for image editing.
  }
  \label{gen1}
\end{figure}

\begin{figure}[t]
  \centering
  \centerline{\includegraphics[width=0.95\linewidth]{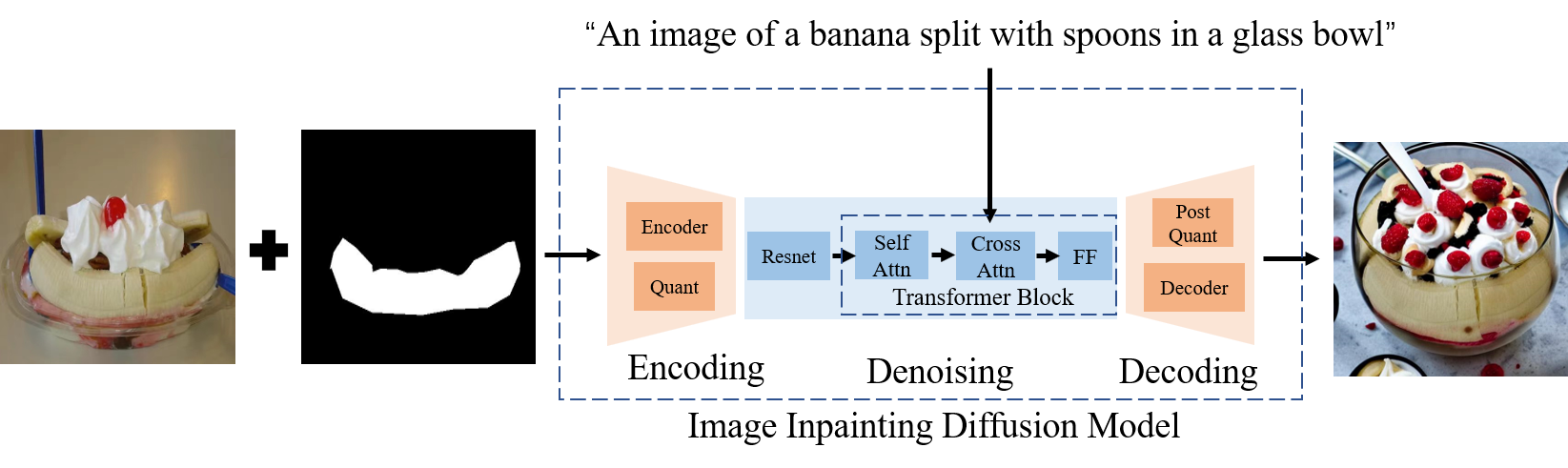}}
  \caption{
  The workflow of image inpainting latent diffusion model for image editing.
  }
  \label{gen2}
\end{figure}

However, diffusion models implement the denoising process on the image level, leading to low efficiency and high computation resource costs. Then, latent diffusion models are proposed to enhance the efficiency by doing the denoising operation on the latent space instead of directly on the images. Specifically, latent diffusion models utilize a variational autoencoder to transform the image between image and latent space. Furthermore, the prompt embedding can be fused into the latent space via the cross attention mechanism in the denoising process for prompt-guided image generation. In addition to image generation, latent diffusion models enable image editing through conditioning. The denoising network initializes the latent feature with a combination of the given image embedding through the encoder and a random sampled noise, while the input prompt is the instruction on how to modify the given image. This modified version of the latent diffusion model is called the image variation model. Another category of the latent diffusion model for image editing is the image inpainting model, which is conditioning on both a given image and a mask of keeping region. The regions outside the mask will be modified by the diffusion models. The workflows of the two categories of image editing are shown in Figure \ref{gen1} and Figure \ref{gen2}. 


\subsection{Adversarial Attacks}

Given the target image $x$ and a DNN model $f(x)$, adversarial attacks aim to find an adversarial sample $x^{adv}$, which can mislead the DNN model, while it is human-imperceptible, i.e., satisfying the constraint $\left\| x - x^{adv} \right\|_{p} < \epsilon$. $\left\| \cdot \right\|_{p}$ represents the $L_p$ norm, and we focus on the $L_\infty$ norm here to align with previous adversarial attack papers \cite{dong2018boosting, lin2019nesterov, zhang2022improving}.

Prevailing adversarial attacks like FGSM \cite{goodfellow2014explaining} usually craft adversarial samples by solving the following constrained maximization problem for attacking an image classifier model with the cross entropy loss J(x,y), where $y$ is the ground truth label for the image $x$:
\begin{equation} \label{eq1}
\nonumber
    \max_{x^{adv}} \ \ J(x^{adv},y) \ \ \ \ s.t. \left\| x - x^{adv} \right\|_{\infty} < \epsilon.
\end{equation}

However, in reality, the model structure and weights are hidden from the users. Therefore, researchers focus on the adversarial attack under the black-box setting. Among different black-box attack methodologies, transfer-based attacks stand out due to their severe security threat to deep learning-based applications in practice. Therefore, we also examine the transfer-based attacks of diffusion models in this paper. %The ability of adversarial samples crafted by a local source model being able to mislead other black-box target models is called transferability. Transfer-based attacks \cite{zhang2022improving, wu2020boosting, wu2021improving} utilize the transferability of adversarial samples. 
Specifically, transfer-based attacks \cite{zhang2023improving, wu2020boosting,zhang2023transferable}  usually craft adversarial samples with a local source model using white-box attack methods and input the generated adversarial samples to mislead the target model. 

\subsection{Adversarial Attacks on Diffusion Models}
Current adversarial attacks on diffusion models mainly focus on the white-box scenario. \cite{zhuang2023pilot} attacks the text encoder by adding extra meaningless characters to the input prompt to mislead the text encoder for attacking diffusion models. \cite{salman2023raising} tries to attack the image encoder and distort the output image to mislead the functionality of latent diffusion models. However, none of the previous works explore the core of the diffusion models (denoising process). Besides, current research only considers the white-box situation and neglects the black-box setting, which is closer to the real world. In this paper, we go deep into the latent diffusion models to analyze the vulnerability of the modules inside diffusion models under the white-box scenario. We also examine the attacking performance under the transfer-based black-box setting from the perspectives of both the prompt-transfer and the model-transfer.

\section{Methodology}

\subsection{Problem Statement}

We evaluate the robustness of latent diffusion models from the perspective of adversarial attacks. Unlike the previous adversarial attacks on image classifier with a clear objective function, it is hard to measure the effectiveness of adversarial attacks on diffusion models. Therefore, we prose to launch feature-level attacks on different components inside diffusion models to perturb the output of each module within the $L_{p}$ constraint. If the representation of the target module is largely destroyed by the adversarial attack, the functionality of the module model is misled. Unlike the adversarial attacks on image classifier to maximize the cross entropy, we aim to distort the feature representation of the attacking module. Thus, we maximize the $l_2$ distance between the intermediate representation of the adversarial example and the original one inside the latent diffusion model. The attacking equation is shown in Equation \ref{eq7}, where $f^m(\cdot)$ represents the output of the module $m$ inside the model $f$ with input image $x$ and prompt $p$.

\begin{equation}
    \max_{x^{adv}} \ \ \left\| f^{m}(x,p) - f^{m}(x^{adv},p) \right\|_{2} \ \ \ \ s.t. \left \| x - x^{adv} \right \|_{\infty} < \epsilon.
    \label{eq7}
\end{equation}

\subsection{Dataset Construction}

\begin{figure}[htbp]
  \centering
  \centerline{\includegraphics[width=0.98\linewidth]{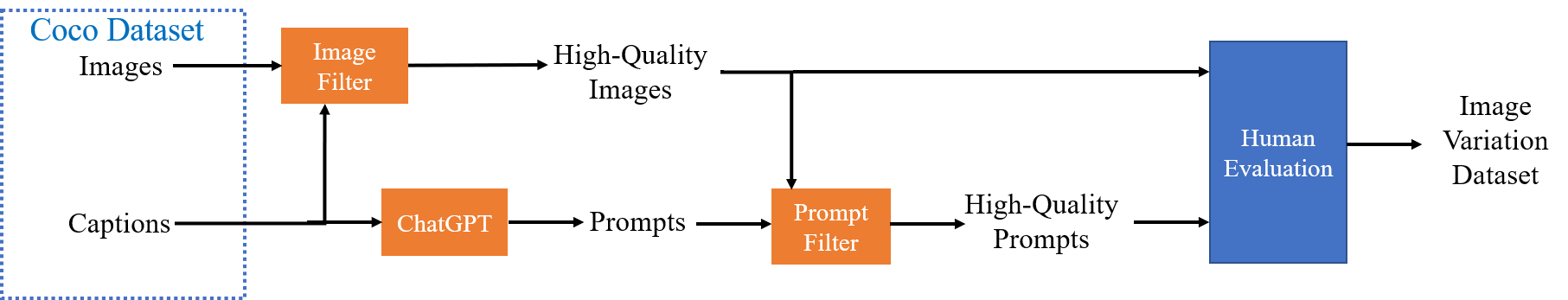}}
  \caption{
  The automatic dataset construction pipeline for image variation diffusion model.
  }
  \label{fig1}
\end{figure}

\begin{figure}[htbp]
  \centering
  \centerline{\includegraphics[width=0.98\linewidth]{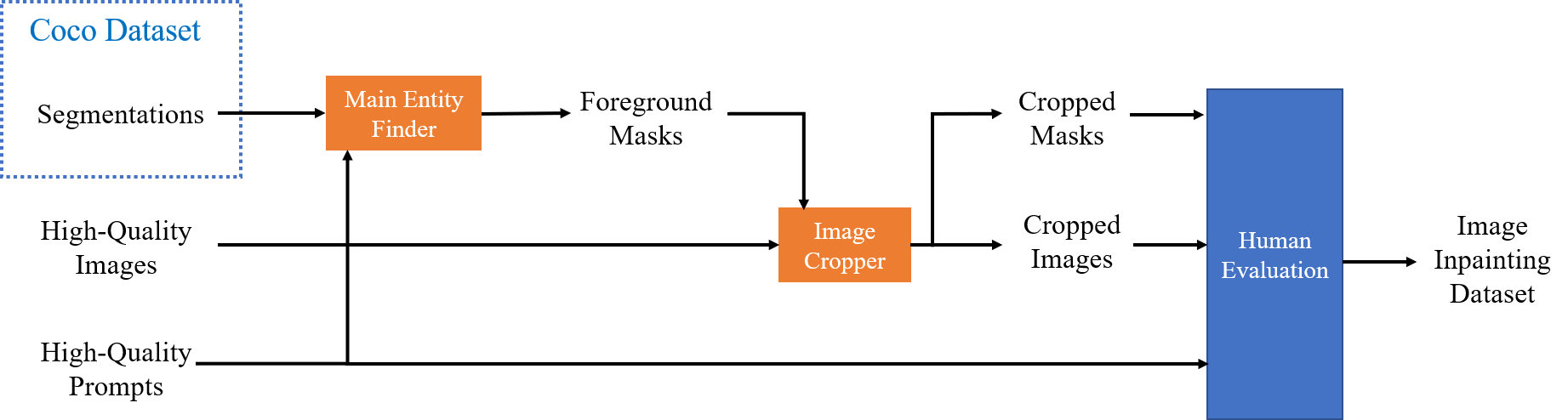}}
  \caption{
  The automatic dataset construction pipeline for image inpainting diffusion model.
  }
  \label{fig2}
\end{figure}

In this section, we present our automatic dataset construction pipeline for building the dataset to analyze the diffusion model robustness. We first state the data source for dataset construction and illustrate the detailed steps to build the dataset. The whole process is automatic, as shown in Figure \ref{fig1} and Figure \ref{fig2}, and human is only required to further guarantee the quality of the dataset at the last step. Please note that our dataset is not only available for evaluating image editing diffusion models. The generated prompts can also be utilized to evaluate the robustness of text2image diffusion models.

\subsubsection{Data Source}
To align with the experimental setting of the diffusion model evaluation, we select the validation set of the coco dataset \cite{lin2014microsoft} as the data source for constructing the dataset. The coco validation dataset contains 5000 images sampled from the natural world with 91 classes, which is complex and diverse enough to test the robustness of diffusion models. Furthermore, coco is the most fundamental dataset for a variety of computer vision tasks, including classification, segmentation, and caption, providing abundant information to fulfill the dataset requirement, e.g., modifying the caption to the image editing prompt or extracting the mask for an object.


\subsubsection{Construction Pipeline}
We propose two different pipelines for the adversarial attack on the image editing functionality of diffusion models based on the model category. One pipeline generates input pairs (image, prompt) for the image variation diffusion model, and another pipeline produces input triplets (image, prompt, mask) for the image inpainting diffusion models. 

\textbf{Pipeline Image variation}

The pipeline generates pairs of images and text prompts. Since the image caption task of the coco dataset consists of 5 human descriptions per image, we can modify each description a little bit to a text prompt without changing the main entity. Therefore, we can directly utilize the images from the coco dataset and change the corresponding captions to form the image variation dataset. 

\textbf{Step 1: Data Preprocessing.} The images from the coco dataset are sampled from the real world, but the contents in the image are complex, and sometimes the main entity is not clear. With the aim of selecting the images with clear main entities, we first rank the images based on the average CLIP score \cite{radford2021learning} between the image and each of the image captions. 10\% of the images are selected to form the dataset. Afterward, we select the top 3 captions for each image based on the CLIP score to guarantee the high-quality of image captions. As a result, we obtain 500 high-quality images and three high-quality captions per image.

\textbf{Step 2: Prompt Generation.} In order to automatically modify the image caption to the text prompt, we deploy the strong power of the large language model \cite{ouyang2022training, brown2020language} for modifying the captions. We query the ChatGPT (GPT 3.5 version) by the following question “Please modify the following sentence <xxx> to generate 5 similar scenes without changing the entities.” to generate 5 text prompts per image caption. For the purpose of successful image editing, we rank and select the top-5 text prompts by the CLIP score between the generated prompt and the output image of Stable Diffusion V1-5 \cite{rombach2022high}. Finally, we obtain five high-quality text prompts per image.

\textbf{Extra Step: Human Evaluation.} A human volunteer is asked to rank the data pair based on the visual performance of the generated images and the coherence between the generated image and the prompt. Finally, the top-100 images are selected, and there are 100*5=500 data pairs for the dataset.
\textbf{Pipeline Image Inpainting}

Image inpainting pipeline generates triplets of image, prompt and mask. The mask covers the region of the main entity, and the diffusion models change the remaining part of the image with the guidance of the prompt. We utilize the bounding box in the detection task to discover the main entity of the image and employ the segmentation task to directly achieve the mask of the main entity. The detailed steps for the dataset construction are followed.

\textbf{Step 1: Data Preprocessing.} The same to pipeline image variation.

\textbf{Step 2: Prompt Generation.} The same to pipeline image variation.

\textbf{Step 3: Main Entity Finder.} We assume the size of the main entity as well as the similarity between the main entity and the image should be large. Based on the first assumption, the top-5 large objects in the image have the potential to be the main entity. Then, We select the object with the highest CLIP score between the image and each category of the potential object to be the main entity. Besides, if other top-5 objects have the same category as the main entity, we consider there are multiple main entities inside the image, and we take the union of their masks.

\textbf{Step 4: Image Cropper.} To avoid the diffusion models being unaware of the main entity, we should guarantee the main entity object takes a large region in the image. Therefore, we adaptively center-crop the image based on the size of the main entity.

\textbf{Extra Step: Human Evaluation.} The same to pipeline image variation.

\subsection{Attacking Modules}

The image editing latent diffusion models contain three crucial processes: encoding, denoising, and decoding as shown in Figure \ref{gen1} and Figure \ref{gen2}. The encoding process first utilizes a variational autoencoder to encode the image and a vector quantization layer to regularize the latent space. We select the image encoder and the quantization layer as the attacking modules for the encoding part. Then, the Unet denoises the latent space and combines the information from the prompt step by step. In each step, the Unet consists of two basic components: resnet block and transformer block, which computes the latent features and combines the information from the prompt, respectively. Especially, the transformer block includes self-attention, cross-attention, and feed-forward layers. Therefore, we select Resnet, self-attention, cross-attention, and feed-forward layers (FF) for adversarial attacks. Finally, the latent features are decoded to the output image through a post-quantization layer and the decoder part of the variational autoencoder, so we choose to attack post-quantization and the decoder.

\section{Experiment}

In this section, we conduct extensive experiments to analyze the robustness of latent diffusion models under different experiment settings. We first specify the setup of the experiments. Then, we present the white-box attacking results under different experiment settings. Besides, we demonstrate the black-box attacking performance under two transfer settings. Finally, we illustrate the robustness of the adversarial examples. 

\subsection{Experimental Setup}

\textbf{Target Model}. We consider both image variation and image inpainting models as the target models. For image variation models, we choose four widely used models, containing Stable Diffusion V1-4 (SD-v1-4) \cite{rombach2022high}, Stable Diffusion V1-5 (SD-v1-5), Stable Diffusion V2-1 (SD-v2-1), and Instruct-pix2pix (Instruct) \cite{brooks2022instructpix2pix}. Different versions of diffusion models have the same structure. The higher version is further trained based on the previous version. While, Instruct-pix2pix has a different structure with Stable Diffusion Models. For image inpainting models, we consider two models: Stable Diffusion v1-5 and Stable Diffusion v2-1. Besides, we include three input-level defense methods that are robust against adversarial attacks. These defenses cover random resizing and padding (R\&P) \cite{xie2017mitigating}, JPEG compression \cite{liu2019feature}, and Gaussian noise (Gaussian) \cite{li2019certified}.

\begin{table}
\caption{The attacking performance against different modules inside SD-v1-5 image variation model by various measurement. The best results are marked in bold.}
\centering
\setlength{\tabcolsep}{2.0mm}{
\begin{tabular}{|c|cccccc|} 
\hline
Module & CLIP & PSNR & SSIM & MSSSIM & FID & IS \\ 
\hline
Encoder & 33.82 & 15.58 & 0.226 & 0.485 & 172.5& 15.05\\
Quant & 34.54 & 16.19 & 0.250 & 0.533& 168.2& 15.77\\
Resnet & \bf 29.89 & \bf 11.82 & \bf 0.076 & \bf 0.270 & \bf 206.3& 16.93\\
Self Attn & 32.37 & 14.91 & 0.108 & 0.305& 206.2& \bf 12.19\\
Cross Attn & 34.48 & 16.22 & 0.250 & 0.557& 171.3& 15.24\\
FF & 31.72 & 13.49 & 0.096 & 0.290& 190.1& 17.33\\
Post Quant & 33.17 & 13.39 & 0.169 & 0.402& 202.8& 17.0\\
Decoder & 34.14 & 15.05 & 0.223 & 0.509& 184.0& 20.59\\
\hline
Gaussian & 34.18 & 15.49 & 0.198 &0.510 & 179.0& 15.41\\
Benign & 34.74 & $\infty$ & 1.000 & 1.000 &167.9 &19.86 \\
\hline
\end{tabular}}
\label{table1}
\end{table}

\textbf{Metric}. We evaluate the performance of adversarial attacks on diffusion models from two perspectives: the quality of the generated image and the functionality of image editing. The quality of generated image are measured via Fréchet Inception Distance (FID) \cite{heusel2017gans} and Inception Score (IS) \cite{salimans2016improved}. The evaluations of image editing functionality are categorized into two angles: normal function disruption and expected function disruption. The normal function disruption measures how the function is influenced by the adversarial examples compared with benign input. We measure the normal function disruption through the similarity of generated images between the adversarial examples and benign ones. We select Peak-Signal-to-Noise Ratio (PSNR), Structural Similarity Index Measure (SSIM) \cite{wang2004image}, and Multi-Scale Structural Similarity Index Measure (MSSSIM) \cite{wang2003multiscale} as the similarity metrics. The expected function disruption measures how the function is influenced by the adversarial examples compared with the expected output. The expected function disruption is measured via the similarity between the generated image and text prompt by the CLIP score \cite{radford2021learning}.

\textbf{Parameter}. All the experiments are conducted on an A100 GPU server with a fixed random seed. Following \cite{kurakin2016adversarial}, we set the maximum perturbation budget $\epsilon = 0.1$, the number of attack iterations $T = 15$, and the step length $\epsilon' = 0.01$. The number of the diffusion step is set to be 15 for attack and 100 for inference. We further take strength and guidance of the diffusion models to be 0.7 and 7.5 by default.

\subsection{White-box Performance}

%\begin{table}
%\centering
%\setlength{\tabcolsep}{1.0mm}{
%{
%\begin{tabular}{|c|cccccc|} 
%\hline
%Module & CLIP & PSNR & SSIM & MSSSIM & FID & IS \\ 
%\hline
%Encoder & 33.04 & -33.80 & 0.255 & 0.538& 174.8& 17.69\\
%Quant & 34.05 & -33.3 & 0.271 &0.572 &155.6 &24.25 \\
%Resnet & 33.44 & -34.6 & 0.219 & 0.495& 172.3& 14.82\\
%Self Attn & 33.61 & -34.24 & 0.220 &0.500 & 162.2& 18.99\\
%Cross Attn & 33.50 & -33.31 & 0.275 &0.572 & 154.6& 23.93\\
%FF & 33.57 & -33.86 & 0.221 & 0.500& 162.4&20.97 \\
%Post Quant & 33.69 & -34.84 & 0.248 &0.520 & 166.3& 21.65\\
%Decoder & 34.17 & -33.06 & 0.282 & 0.594& 155.7& 24.73\\
%\hline
%Gaussian & 34.31 & -32.18 & 0.316 & 0.628& 161.1& 19.78\\
%Benign & 34.4 & $\infty$ & 1.000 & 1.000& 156.4& 22.45\\
%\hline
%\end{tabular}}
%\caption{The attacking performance against different modules inside SD-v1-5 image inpainting by various measurement. The best results are marked in bold.}
%\label{table2}
%\end{table}

\begin{table}
\caption{The white-box attacking performance against different image variation diffusion models by various measurement. The best results are marked in bold.}
\centering
\setlength{\tabcolsep}{2.0mm}{
\begin{tabular}{|c|c|cccccc|} 
\hline
Model & Module & CLIP & PSNR & SSIM & MSSSIM & FID & IS \\ 
\hline
\multirow{5}{*}{SD-v1-4} & Encoding & 34.09 & 15.47 & 0.225 & 0.485 & 172.7& 15.8\\
 & Unet & \bf 30.93 & \bf 12.23 & \bf 0.080 & \bf 0.288& \bf 193.4& 19.8\\
 & Decoding & 33.00 & 13.30 & 0.168 & 0.405& 190.1& \bf 14.5\\
 & Gaussian & 34.05 & 15.42 & 0.199 & 0.509& 174.4& 19.8\\
 & Benign & 34.54 & $\infty$ & 1.000 & 1.000  & 174.5& 19.8\\
\hline
\multirow{5}{*}{SD-v1-5} & Encoding & 33.82 & 15.58 & 0.226 & 0.485 & 172.5& \bf 15.05\\
 & Unet & \bf 29.89 & \bf 11.82 & \bf 0.076 & \bf 0.270 & \bf 206.3& 16.93\\
 & Decoding & 33.17 & 13.39 & 0.169 & 0.402& 202.8& 17.00\\
 & Gaussian & 34.18 & 15.49 & 0.198 &0.510 & 179.0& 15.41\\
 & Benign &  34.74 & $\infty$ & 1.000 & 1.000 &167.9 &19.86 \\
\hline
\multirow{5}{*}{SD-v2-1} & Encoding & 30.62 & 14.13 & 0.156 & 0.392 & 217.8& \bf 10.39\\
 & Unet & 27.98 & \bf 12.25 & \bf 0.111 &\bf 0.314 &\bf 217.4 &13.57 \\
 & Decoding & \bf 27.61 & 12.30 & \bf 0.111 & 0.317 & 212.2& 13.93\\
 & Gaussian & 31.77 & 13.70 & 0.156 & 0.425& 202.6& 14.68\\
 & Benign & 32.03 & $\infty$ & 1.000 & 1.000  & 201.8& 13.74\\
\hline
\multirow{5}{*}{Instruct} & Encoding & 33.94 & 9.89 & \bf 0.125 & \bf 0.338& \bf 180.9& \bf 16.06\\
 & Unet & \bf 31.65 & 11.87 & 0.153 & 0.399& 175.8& 17.46\\
 & Decoding & 33.75 & \bf 9.81 & 0.146 & 0.364& 167.9& 20.86\\
 & Gaussian & 33.98 & 12.13 & 0.197 &0.457 & 167.9& 20.86\\
 & Benign & 34.45 & $\infty$ & 1.000 & 1.000  & 153.75 & 22.89\\
\hline 
\end{tabular}}
\label{table3}
\end{table}

%We also analyze the attacking performance on different modules of SD-v1-5 Inpainting model as shown in Table \ref{table2}. Compared with attacking image variation model, it is challenging to mislead the functionality since the output image keeps the main entity and the editing region is limited. However, the adversarial attacks can still reduce the CLIP score by more than 1.3, and the image quality as well as the normal function are influenced. The qualitative results are shown in Figure []. Among the modules, attacking the Resnet achieves the best performance on destroying normal function and image quality, but it is a little bit inferior than attacking the encoder. We think image inpainting model has the mask input and the edited region is limited, so attacking the Unet may overfit to the model leading to a slightly lower expected function disruption and a higher normal function disruption compared with attacking encoder. Therefore, Resnet and encoder are more vulnerable than other modules.

\begin{figure}[htbp]
\ \ \ \ \ \ \ \ \ \ Clean\ \ \ \ \ \ \ \ \ \ \ \ \ \ \ \ \ Gaussian \ \ \ \ \ \ \ \ \ \ \ \ \ \ \ Encoding\ \ \ \ \ \ \ \ \ \ \ \ \ \ \ \ \ Unet\ \ \ \ \ \ \ \ \ \ \ \ \ \ \ \ \ Decoding \\
\centering
\begin{minipage}[t]{0.18\textwidth}
\centering
\includegraphics[width=2.5cm]{image/ori.png}
\end{minipage}
\begin{minipage}[t]{0.18\textwidth}
\centering
\includegraphics[width=2.5cm]{image/gauss.png}
\end{minipage}
\begin{minipage}[t]{0.18\textwidth}
\centering
\includegraphics[width=2.5cm]{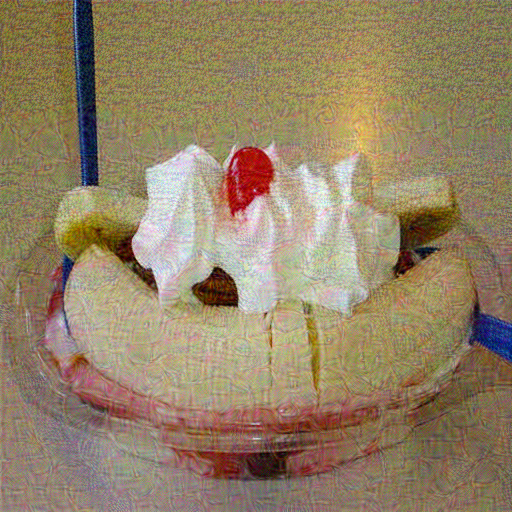}
\end{minipage}
\begin{minipage}[t]{0.18\textwidth}
\centering
\includegraphics[width=2.5cm]{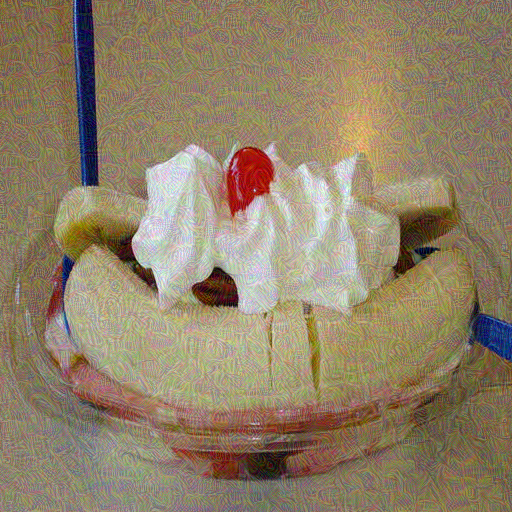}
\end{minipage}
\begin{minipage}[t]{0.18\textwidth}
\centering
\includegraphics[width=2.5cm]{image/postquant.png}
\end{minipage}
Adversarial Images \\
\begin{minipage}[t]{0.18\textwidth}
\centering
\includegraphics[width=2.5cm]{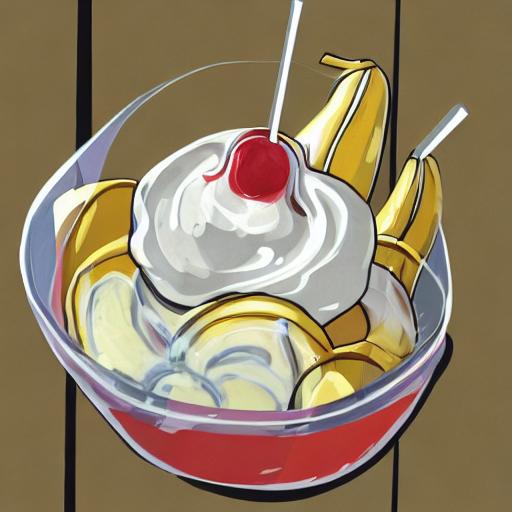}
\end{minipage}
\begin{minipage}[t]{0.18\textwidth}
\centering
\includegraphics[width=2.5cm]{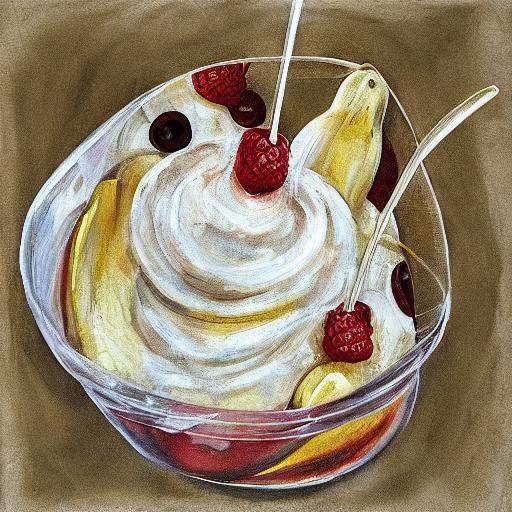}
\end{minipage}
\begin{minipage}[t]{0.18\textwidth}
\centering
\includegraphics[width=2.5cm]{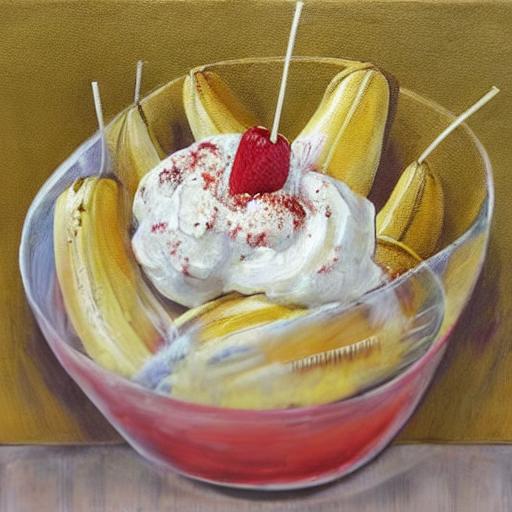}
\end{minipage}
\begin{minipage}[t]{0.18\textwidth}
\centering
\includegraphics[width=2.5cm]{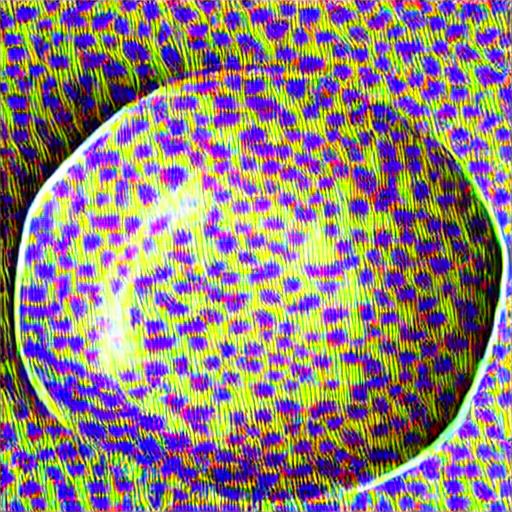}
\end{minipage}
\begin{minipage}[t]{0.18\textwidth}
\centering
\includegraphics[width=2.5cm]{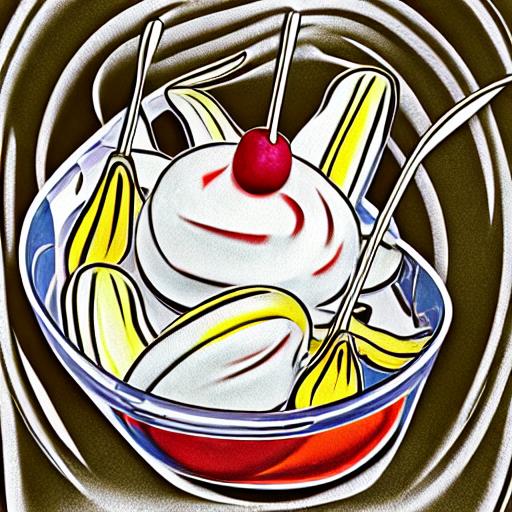}
\end{minipage}
Generated Images\\
\caption{Visualization of adversarial images and their output of the SD-v1-5 image variation model with the prompt "An artwork of a banana split in a plastic bowl".}
\label{vis}
\end{figure}

We first illustrate the white-box attacking performance on different modules inside the Stable Diffusion V1-5 image variation and inpainting models, respectively. Then we select the most vulnerable module for each process and evaluate the white-box performance on other diffusion models.

As shown in Table \ref{table1}, we analyze the attacking performance on different modules of the SD-v1-5 image variation model. We first analyze the expected function that the CLIP scores of adversarial attacks are all lower than the Gaussian noise, validating that the diffusion models are susceptible to adversarial attacks. In addition, attacking the denoising process achieves the lowest CLIP score compared with attacking the encoding or decoding process. Astonishingly, the CLIP score is reduced to 29.89 with a marginal drop of 4.85, destroying the expected function severely. From the point of view of both normal function and image quality, attacking the denoising process and especially the Resnet module outperforms attacking other processes and modules. We also observe that attacking the cross attention performs badly. We think the reason is the prompt information dominates the cross attention module, so attacking on the image is not effective, which requires textual adversarial attacks to corrupt this module. The qualitative results are shown in Figure \ref{vis}. In general, the generated images by the adversarial examples are of low visual quality, and the editing is meaningless. Furthermore, the qualitative results also validate our conclusion that attacking the denoising process is effective. We can achieve a similar conclusion on attacking image inpainting diffusion models in Appendix.

In addition to the white-box analysis on Stable Diffusion v1-5, we illustrate the attacking performance on other diffusion models as shown in Table \ref{table3}. We select to attack the most vulnerable modules of each process and measure the attacking performance. We can see all the diffusion models are susceptible to adversarial attacks, which can largely influence the normal function and image quality. The Instruct-pix2pix is more robust compared with standard stable diffusion models with higher CLIP score and similarity with the benign output. Attacking the Unet is consistently effective compared with attacking other processes of the image variation diffusion models. More white-box performance of the image inpainting model is shown in the Appendix.

\begin{table}
\caption{The prompt-transfer attacking performance on SD-v1-5 image variation model by various measurement. The best results are marked in bold.}
\centering
\setlength{\tabcolsep}{2.0mm}{
\begin{tabular}{|c|cccccc|} 
\hline
Module & CLIP & PSNR & SSIM & MSSSIM & FID & IS \\ 
\hline
Encoding & 33.82 & 15.58 & 0.226 & 0.485 & 172.5& \bf 15.05\\
Unet & \bf 28.27 & \bf 11.92 & \bf 0.074 & \bf 0.271 & 183.33& 24.00\\
Decoding & 30.18 & 13.55 & 0.166 & 0.400 & \bf 187.85& 20.56\\
\hline
Gaussian & 34.18 & 15.49 & 0.198 &0.510 & 179.0& 15.41\\
\hline
\end{tabular}}
\label{table4}
\end{table}

\begin{table}
\caption{The model-transfer attacking performance against different image variation diffusion models. The best results are marked in bold.}
\centering
\setlength{\tabcolsep}{2.0mm}{
\begin{tabular}{|c|c|cccc|} 
\hline
Model & Module & SD-v1-4 & SD-v1-5 & SD-v2-1 & Instruct \\ 
\hline
\multirow{3}{*}{SD-v1-4} & Encoding & 34.09 & 33.82 & 30.60 & 33.94\\
 & Unet & \bf 30.93 & \bf 31.40 & \bf 29.13 & \bf 33.70\\
 & Decoding & 33.00 & 33.07 & 29.71& 33.86\\

\hline
\multirow{3}{*}{SD-v1-5} & Encoding & 34.11 & 33.82 & 30.65& 33.94\\
 & Unet & \bf 29.73 &\bf 29.89 & \bf 28.48& \bf 33.30\\
 & Decoding & 33.15 & 33.17 & 29.47& 33.93\\

\hline
\multirow{3}{*}{SD-v2-1} & Encoding & 34.09 & \bf 33.84 & 30.62 & 33.93\\
 & Unet & \bf 33.85 & 33.94 & 27.98 & \bf 33.90\\
 & Decoding & 33.88 & 34.07 & \bf 27.61 & 34.03\\

\hline
\multirow{3}{*}{Instruct} & Encoding & \bf 33.96 & 34.07 & 31.24 & 33.94 \\
 & Unet & 34.08 & \bf 34.02 & \bf 30.52 & \bf 31.65\\
 & Decoding & 34.23 & 34.22 & 31.74 & 33.75\\
 
\hline 
\end{tabular}}
\label{table5}
\end{table}

\begin{figure}[htbp]
\ \ \ \ \ \ \ SD-v1-4\ \ \ \ \ \ \ \ \ \ \ \ \ \ \ \ \ \ \ \ \ \ \ \ SD-v2-1 \ \ \ \ \ \ \ \ \ \ \ \ \ \ \ \ \ \ \ \ \ \ \ \ R \& P\ \ \ \ \ \ \ \ \ \ \ \ \ \ \ \ \ \ \ \ \ \ \ \ \ \ \ \ JPEG \\
\centering
\begin{minipage}[t]{0.23\textwidth}
\centering
\includegraphics[width=2.5cm]{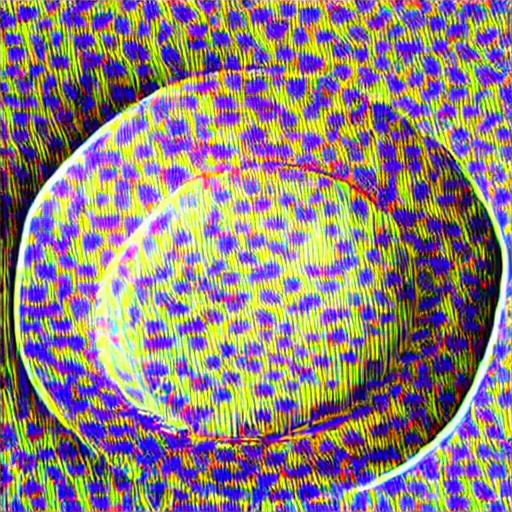}
\end{minipage}
\begin{minipage}[t]{0.23\textwidth}
\centering
\includegraphics[width=2.5cm]{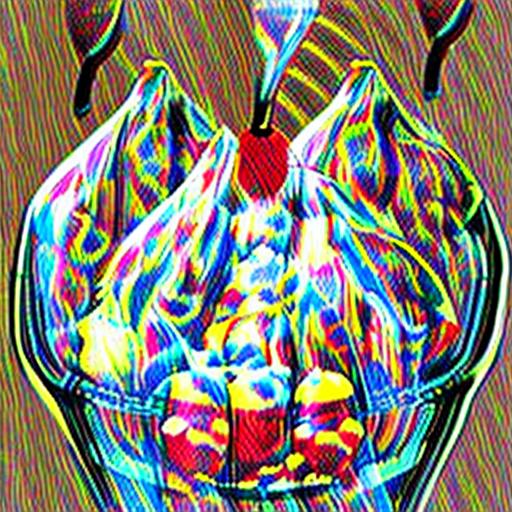}
\end{minipage}
\begin{minipage}[t]{0.23\textwidth}
\centering
\includegraphics[width=2.5cm]{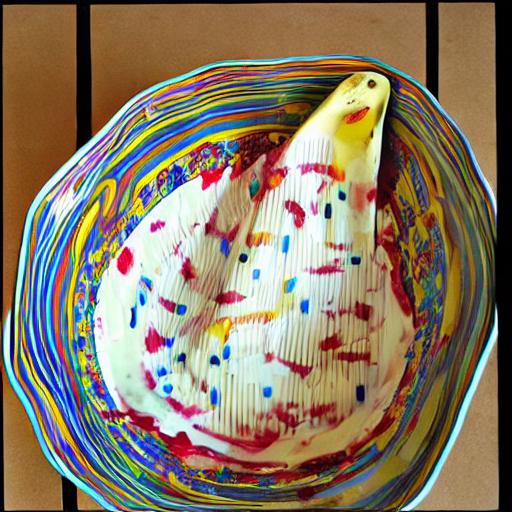}
\end{minipage}
\begin{minipage}[t]{0.23\textwidth}
\centering
\includegraphics[width=2.5cm]{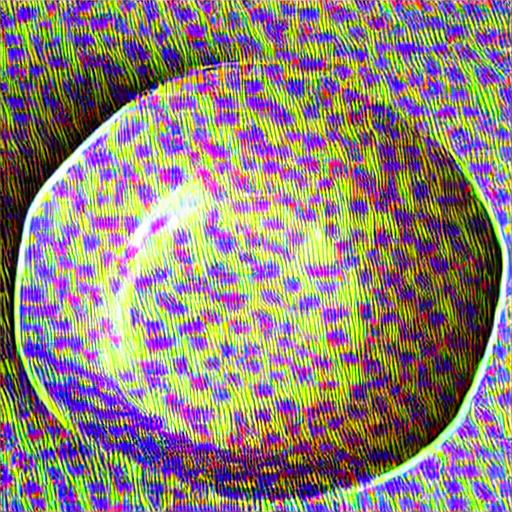}
\end{minipage}
\caption{Visualization of generated images by model-transfer attacks on SD-v1-4 and SD-v2-1, and generated images of the SD-v1-5 against defense mechanisms with the prompt "An artwork of a banana split in a plastic bowl".}
\label{vis2}
\end{figure}

\subsection{Black-box Performance}

In this section, we demonstrate the black-box performance, especially, the transfer-based attacks. Unlike the transfer-based attacks for image classifiers, we categorize the transfer-based attacks for diffusion models into two classes: model-transfer and prompt-transfer. Model-transfer is similar to transfer-based attacks on image classifiers in that the generated adversarial examples by one diffusion model can mislead the other diffusion models. Prompt-transfer is from the perspective of data that the generated adversarial examples can also influence the diffusion model with other similar prompts.

We first illustrate the prompt-transfer performance on the Stable Diffusion V1-5 image variation model. We craft adversarial image on one prompt and transfer the image to other prompts to evaluate the generative results. As shown in Table \ref{table4}, the prompt-transfer achieves similar results on CLIP score and normal function with the original prompt. The adversarial images crafted by one prompt can also mislead the guidance of other similar prompts. Therefore we can successfully mislead the diffusion models by adversarial samples without knowing the exact text prompts.

Then, we consider the model-transfer performance that we craft adversarial examples on the victim model and directly transfer the adversarial samples to the other models. As shown in Table \ref{table5}, the transfer attacks can mislead the expected function because the CLIP scores are all reduced compared with the benign results in Table \ref{table3}. Instruct-pix2pix is hard to transfer to other standard diffusion models, and other diffusion models are also hard to transfer to Instruct-pix2pix because of the different model architectures. Considering the transfer attacks between different versions of standard stable diffusion models, adversarial examples crafted by SD-v1-4 and SD-v1-5 are easy to mislead SD-v2-1. The experiment result means the SD-v2-1 is more vulnerable, and the defects inside the previous versions of stable diffusion models are inherited by SD-v2-1. Furthermore, adversarial samples from SD-v2-1 cannot largely mislead SD-v1-4 or SD-v1-5, representing the deficiencies of SD-v2-1 are not the common problems for SD-v1-4 or SD-v1-5. Combining the two observations, we conclude the problems of SD-v1 are inherited by SD-v2, and SD-v2 has more defects compared with SD-v1. With the updating of the stable diffusion models, models become more vulnerable and have more defects, which raises the alarm about the robustness of diffusion models. The researchers should also consider the robustness issue when they update the version of diffusion models. The qualitative results are shown in Figure \ref{vis2}. We can see that transfer-based attacks can mislead the functionality of other diffusion models, raising concerns about the robustness of diffusion models. More model-transfer results on image inpainting diffusion models are shown in Appendix.

\begin{table}
\caption{The attacking performance against SD-v1-5 image variation model by various defense mechanisms. The best results are marked in bold.}
\centering
\setlength{\tabcolsep}{2.0mm}{
\begin{tabular}{|c|c|cccccc|} 
\hline
Methods & Module & CLIP & PSNR & SSIM & MSSSIM & FID & IS \\ 
\hline
\multirow{3}{*}{R \& P} & Encoding & 33.94 & 12.42 & 0.187 &0.413 & 179.0& 24.36\\
 & Unet & \bf 33.84 & \bf 12.20 & \bf 0.173 & \bf 0.397& \bf 181.0& \bf 23.64\\
 & Decoding & 33.92 & 12.30 & 0.190 & 0.418& 177.8& 24.90\\
\hline
\multirow{3}{*}{JPEG} & Encoding & 34.11 & 15.94 & 0.252 & 0.513& 196.0& 16.02\\
 & Unet & \bf 30.75 & \bf 12.23 & \bf 0.087 &\bf 0.299 &\bf 234.8 & \bf 9.74\\
 & Decoding & 33.50 & 14.00 & 0.195 & 0.437& 189.6& 19.02\\
\hline
\multirow{3}{*}{Gaussian} & Encoding & 33.84 & 14.58 & 0.140 &0.441 & 170.3& \bf 22.04\\
 & Unet & \bf 30.75 & \bf 12.06 & \bf 0.072 & \bf 0.289 & \bf 183.1& 22.69\\
 & Decoding & 33.39 & 13.50 & 0.127 & 0.401& 173.4& 22.94\\
\hline

\hline 
\end{tabular}}
\label{table6}
\end{table}

\subsection{Robustness of Adversarial Examples}

We further analyze the robustness of adversarial examples because the robustness of the adversarial examples has practical issues. If the adversarial images are robust, we can deploy them to avoid the image editing model's misuse in real life. We consider three kinds of defense mechanisms on the image to mitigate the adversarial influence, including geometry transformation, compression, and noise. Especially, We select Random Resizing and Padding (R \& P), JPEG compression (JPEG), and Gaussian noise (Gaussian) to evaluate the robustness of adversarial examples. All the defense mechanisms can mitigate the adversarial influence on the normal function and expected function as shown in Table \ref{table6}. Significantly, the geometry transformation (R \& P) can largely alleviate the adversarial problems with a 3.95 increment of the CLIP score on attacking the Unet. The qualitative results are also shown in Figure \ref{vis2}. Although the defense mechanisms can mitigate the adversarial issues, the expected function is still far from satisfaction in the qualitative examples.

\section{Conclusion}

In this paper, we explore the robustness of the diffusion models from the perspective of adversarial attacks. We first demonstrate that the denoising process, especially the Resnet module is the most vulnerable component in the diffusion models. We also find Instruct-pix2pix is more robust than stable diffusion models by comparing the white-box performance. We further consider two transfer-based black-box scenarios: prompt-transfer and model-transfer. The adversarial images can transfer well between prompts and models. Significantly, we figure out that adversarial examples from SD-v1 can transfer well to SD-v2, but adversarial samples from SD-v2 cannot transfer well to SD-v1. This phenomenon shows SD-v2 is more vulnerable than SD-v1, and defects inside SD-v1 are inherited by SD-v2. This observation raises concerns about the robustness of the diffusion model development. Besides, geometry transformation can largely mitigate the effect of adversarial examples, but adversarial examples can still mislead the functionality of diffusion models under several defense mechanisms. The discussion of the negative social impact and limitations are attached in the Appendix.

\bibliography{neurips_2023}
\bibliographystyle{neurips_2023}
%%%%%%%%%%%%%%%%%%%%%%%%%%%%%%%%%%%%%%%%%%%%%%%%%%%%%%%%%%%%

\newpage
\section*{Appendix}
\appendix

In this appendix, Section \ref{a} shows the detailed experimental setting for each experiment in the paper. Section \ref{b} illustrates the constructed dataset. Section \ref{c} reports the attacking performance against latent diffusion image inpainting models under the white-box and black-box scenarios. Section \ref{d} demonstrates more qualitative results on different attacking scenarios. Broader impacts and limitations are discussed in Section \ref{e}.

\section{Experimental Setting}
\label{a}

In this section, we further clarify the experimental settings for each experiment in the paper. 

There are, in total, six metrics to measure the functionality of the latent diffusion models from three aspects. CLIP score measures the similarity between the generated image and the edited prompt representing the expected function. A low CLIP score means the generated image is far from the expected editing of the prompt. The normal function disruption is measured by three metrics: PSNR, SSIM, and MSSSIM. They compute the similarity between the benign input and the corresponding adversarial input, and a low score reflects a large normal function disruption. The quality of the generated image is measured by FID and IS. A lower FID and a higher IS indicate that the generated images are of higher quality.

We select the most well-known latent diffusion models: Stable Diffusion v1-5 for the white-box component experiments. We consider eight different modules inside the SD-v1-5, as explained in Section 3.2 of the paper. We deploy the adversarial attacks described in Equation (2) on the target model's output to destroy the intermediate features of the target module. Since there are multiple denoising steps, we consider to destroy the target module inside the denoising process on all the steps. We also select Gaussian noises as the standard baseline. The experimental results show that the adversarial noises are better than the Gaussian noise.

For the white-box experiment on other latent diffusion models, We select to attack the most vulnerable modules of each process. Specifically, We choose the Encoder module for the Encoding process, Resnet for Unet, and Post Quantization layers for the Decoding process.

We consider two kinds of transfer-based attack scenarios in diffusion attacks. The prompt-transfer means that the adversarial image crafted by one prompt can also mislead the functionality of the diffusion model with a similarly edited prompt input. We test the prompt-transfer in a circulation manner in the experiment. Specifically, we craft one adversarial image by the current prompt and test the prompt-transfer by the next prompt of the image when we fix the image and model. We also evaluate the model-transfer and directly transfer the adversarial image by fixing the image and prompt.

\section{Dataset Illustrations}
\label{b}

\begin{figure}[t]
  \centering
  \centerline{\includegraphics[width=0.77\linewidth]{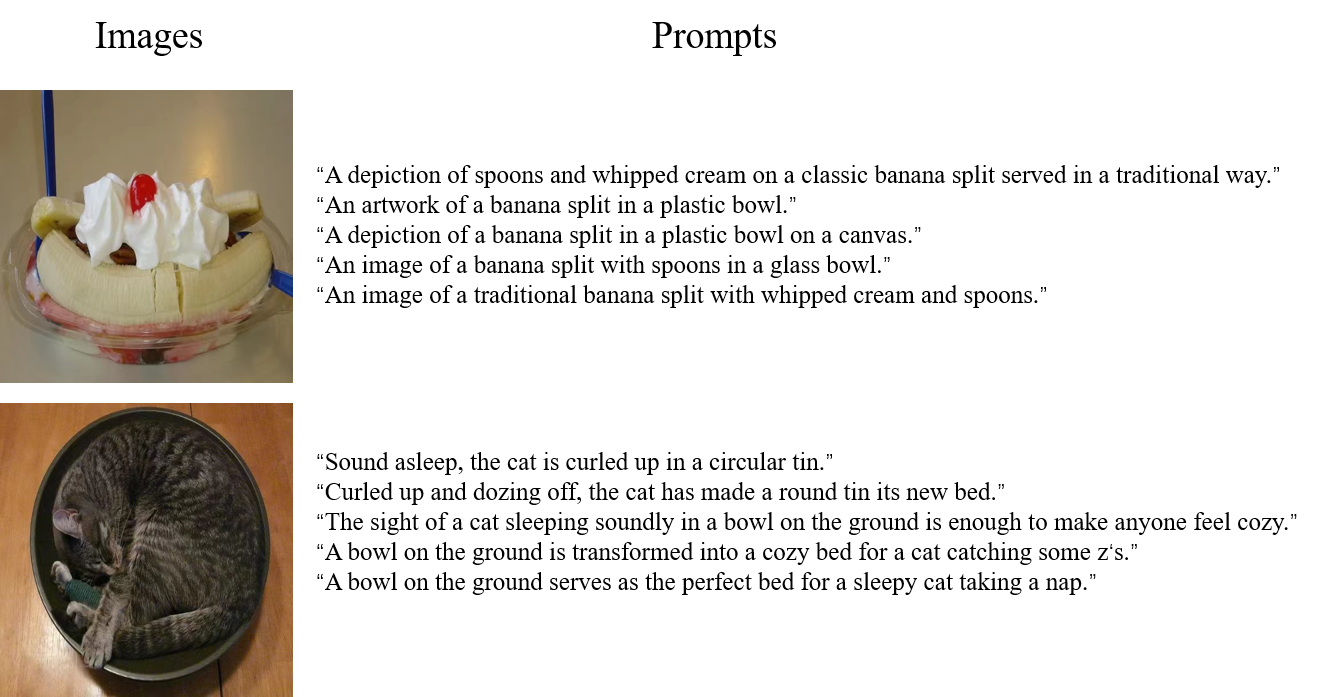}}
  \caption{
  The data pair examples of the constructed dataset for the image variation diffusion models.
  }
  \label{figa1}
\end{figure}

\begin{figure}[t]
  \centering
  \centerline{\includegraphics[width=0.98\linewidth]{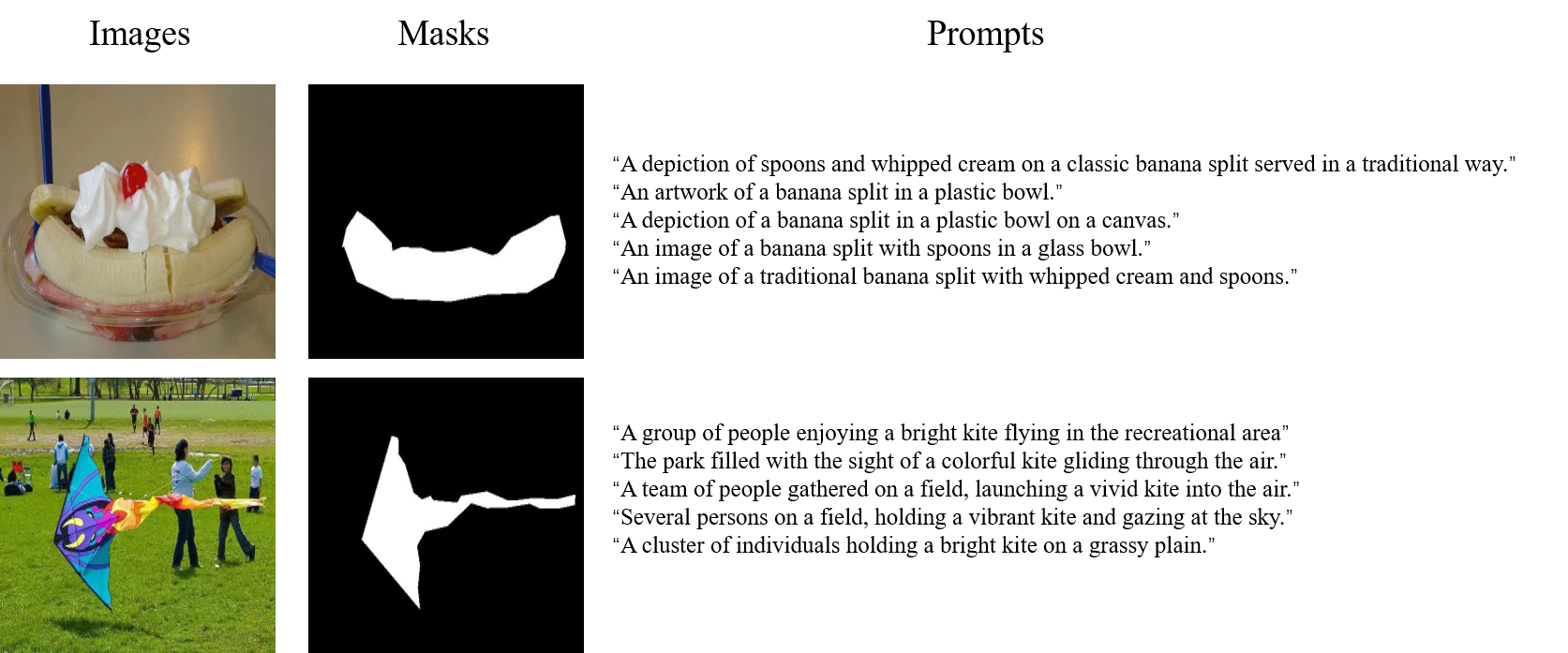}}
  \caption{
  The data triplet examples of the constructed dataset for the image inpainting diffusion models.
  }
  \label{figa2}
\end{figure}

The dataset contains two subsets for different categories of image editing diffusion models. Each subset includes 100 images and five edited prompts per image. The subset for image inpainting diffusion models also has additional 100 image masks for the images. The qualitative results of the dataset are shown in Figure \ref{figa1} and Figure \ref{figa2}. We will release the whole dataset and the code for constructing the dataset to boost the research on the robustness of diffusion models. 


\section{Attacking Performance}
\label{c}
\begin{table} [h]
\caption{The attacking performance against different modules inside the SD-v1-5 image inpainting model. The best results are marked in bold.}
\centering
{
\begin{tabular}{|c|cccccc|} 
\hline
Module & CLIP & PSNR & SSIM & MSSSIM & FID & IS \\ 
\hline
Encoder & \bf 33.04 & 14.33 & 0.255 & 0.538& \bf 174.8& 17.69\\
Quant & 34.05 & 14.83 & 0.271 &0.572 &155.6 &24.25 \\
Resnet & 33.44 & 13.53 & \bf 0.219 & \bf 0.495& 172.3& \bf 14.82\\
Self Attn & 33.61 & 13.89 & 0.220 &0.500 & 162.2& 18.99\\
Cross Attn & 33.50 & 14.82 & 0.275 &0.572 & 154.6& 23.93\\
FF & 33.57 & 14.27 & 0.221 & 0.500& 162.4&20.97 \\
Post Quant & 33.69 & \bf 13.29 & 0.248 &0.520 & 166.3& 21.65\\
Decoder & 34.17 & 15.07 & 0.282 & 0.594& 155.7& 24.73\\
\hline
Gaussian & 34.31 & 15.95 & 0.316 & 0.628& 161.1& 19.78\\
Benign & 34.4 & $\infty$ & 1.000 & 1.000& 156.4& 22.45\\
\hline
\end{tabular}}

\label{tablea2}
\end{table}

We present more results on attacking latent diffusion image inpainting models in this section. We analyze the attacking performance on different modules of the SD-v1-5 image inpainting model and the attacking performance on the selected image inpainting latent diffusion models under the white-box scenario. Besides, we evaluate the model-transfer performance across different versions of the Stable Diffusion models.

\begin{figure}[htbp]
\ \ \ \ \ \ \ \ Clean \ \ \ \ \ \ \ \ \ \ \ \ \ \ \  \ \ Mask \ \ \ \ \ \ \ \ \ \ \ \ \ Generate \ \ \ \ \ \ \ \ \ \ \ \ Encoding\ \ \ \ \ \ \ \ \ \ \ \ \ \ Unet\ \ \ \ \ \ \ \ \ \ \ \ \ \ \ Decoding\ \ \ \ \\
\centering
\begin{minipage}[t]{0.16\textwidth}
\centering
\includegraphics[width=2.2cm]{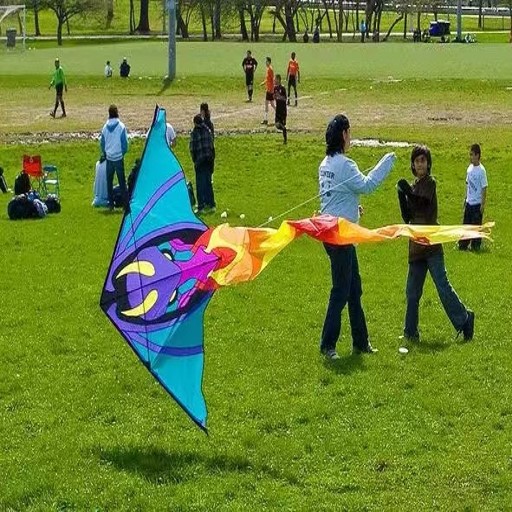}
\end{minipage}
\begin{minipage}[t]{0.16\textwidth}
\centering
\includegraphics[width=2.2cm]{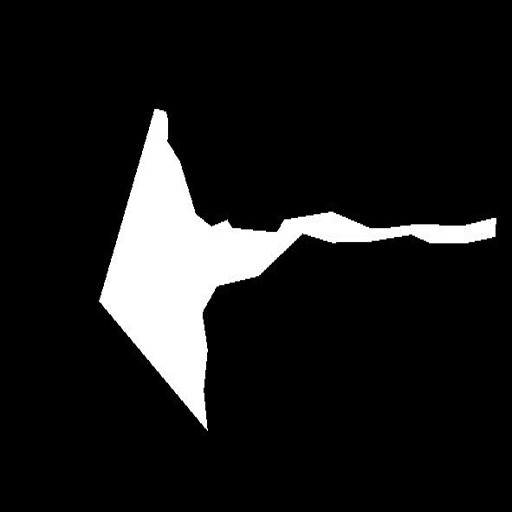}
\end{minipage}
\begin{minipage}[t]{0.16\textwidth}
\centering
\includegraphics[width=2.2cm]{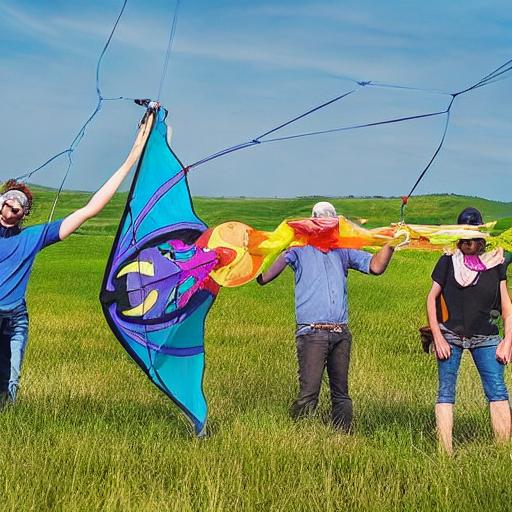}
\end{minipage}
\begin{minipage}[t]{0.16\textwidth}
\centering
\includegraphics[width=2.2cm]{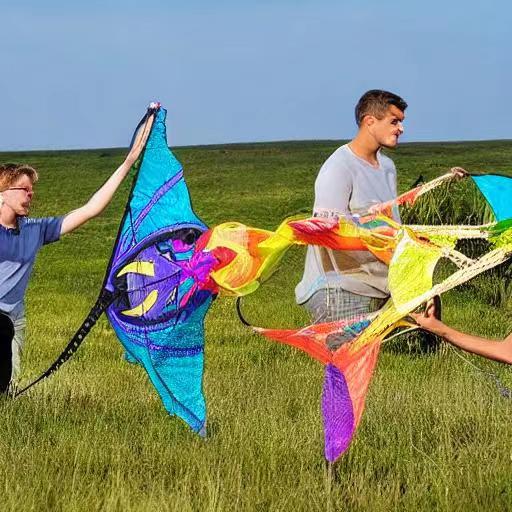}
\end{minipage}
\begin{minipage}[t]{0.16\textwidth}
\centering
\includegraphics[width=2.2cm]{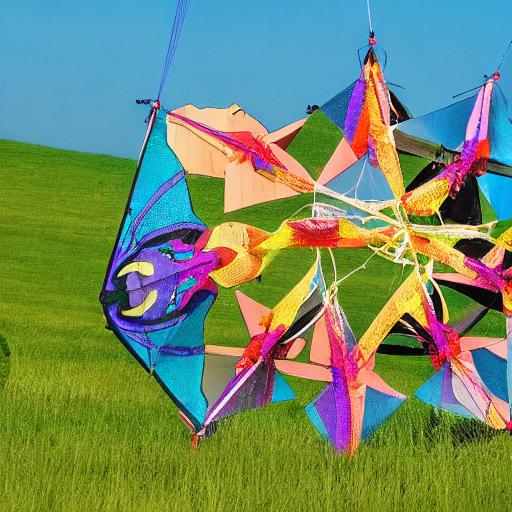}
\end{minipage}
\begin{minipage}[t]{0.16\textwidth}
\centering
\includegraphics[width=2.2cm]{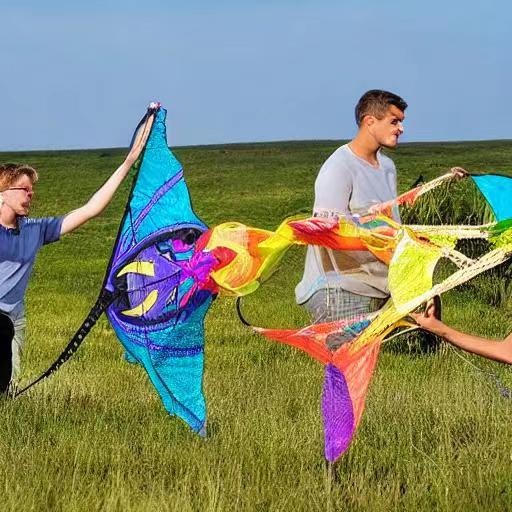}
\end{minipage}
\\"A cluster of individuals holding a bright kite on a grassy plain." \\
\begin{minipage}[t]{0.16\textwidth}
\centering
\includegraphics[width=2.2cm]{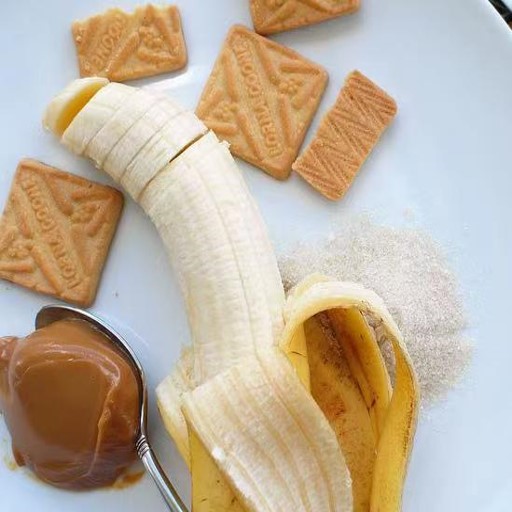}
\end{minipage}
\begin{minipage}[t]{0.16\textwidth}
\centering
\includegraphics[width=2.2cm]{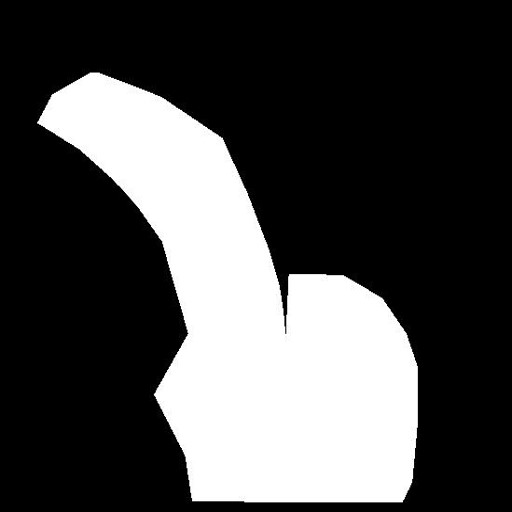}
\end{minipage}
\begin{minipage}[t]{0.16\textwidth}
\centering
\includegraphics[width=2.2cm]{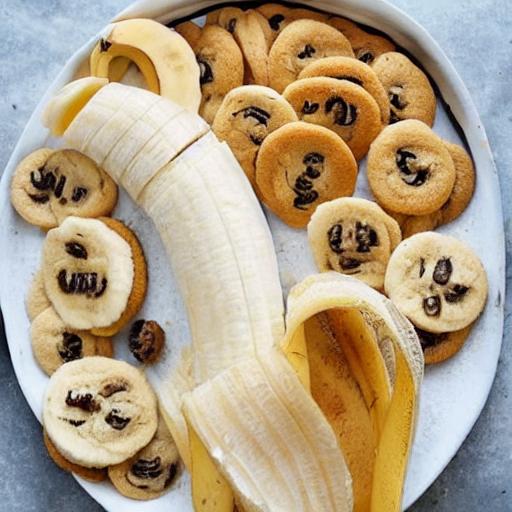}
\end{minipage}
\begin{minipage}[t]{0.16\textwidth}
\centering
\includegraphics[width=2.2cm]{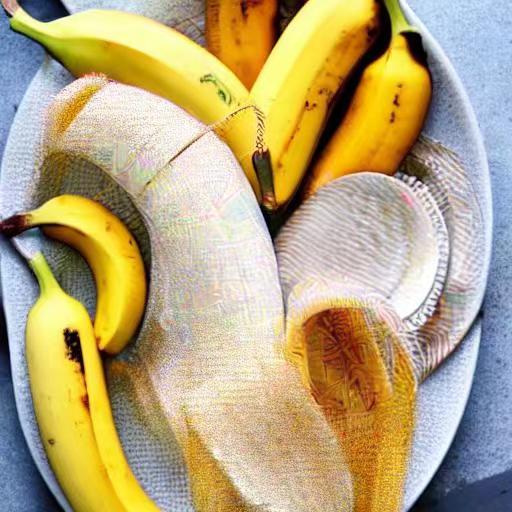}
\end{minipage}
\begin{minipage}[t]{0.16\textwidth}
\centering
\includegraphics[width=2.2cm]{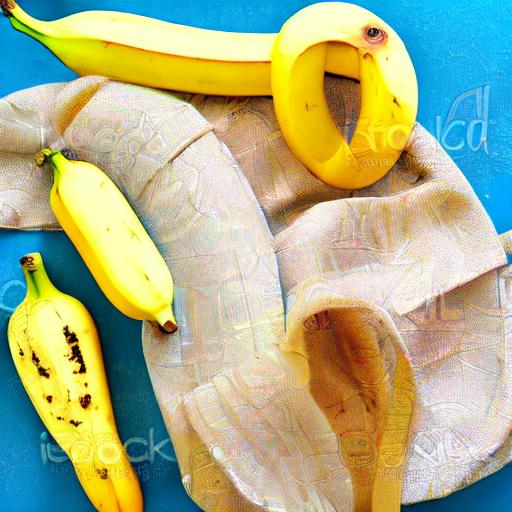}
\end{minipage}
\begin{minipage}[t]{0.16\textwidth}
\centering
\includegraphics[width=2.2cm]{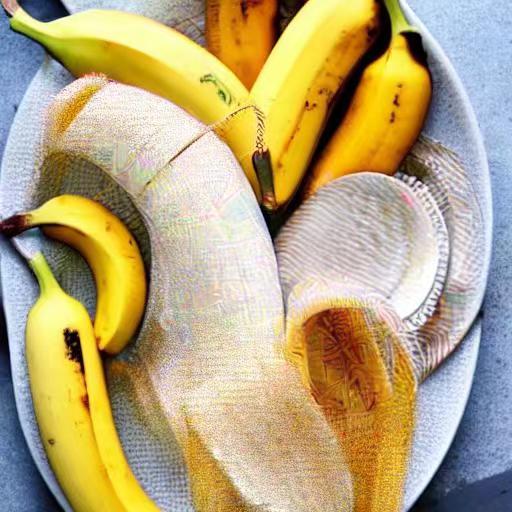}
\end{minipage}
\\"A platter with a banana that is sliced, cookies, and a spoon resting on it." \\
\caption{Visualization of adversarial images and their output of the SD-v1-5 image inpainting model.}
\label{vis_inpainting}
\end{figure}

We show the attacking performance on different modules of the SD-v1-5 inpainting model in Table \ref{tablea2}. Compared with attacking image variation models, it is challenging to mislead its functionality, since the output image keeps the main entity, and the editing region is limited. However, adversarial attacks can still reduce the CLIP score by more than 1.3, and the image quality as well as the normal function are influenced. The qualitative results are shown in Figure \ref{vis_inpainting}. Among the modules, attacking the Resnet achieves the best performance on destroying the normal function and the image quality, but it is a little bit inferior than attacking the encoder. Since an image inpainting model has the mask input, and the edited region is limited, attacking the Unet may overfit the model, leading to a slightly lower expected function disruption and a higher normal function disruption compared with attacking the encoder. Therefore, Resnet and encoder are more vulnerable than other modules for image inpainting diffusion models.

Additionally, as shown in Table \ref{tablea3}, we can draw the same conclusion as the other image inpainting latent diffusion models that the encoding and Unet are vulnerable. Attacking the encoding process can disrupt the expected function, while destroying the Unet is able to mislead the normal function and reduce the image quality, which can be validated from the qualitative results. 

\begin{table}
\caption{The white-box attacking performance against different image inpainting diffusion models. The best results are marked in bold.}
\centering
\setlength{\tabcolsep}{2.0mm}{
\begin{tabular}{|c|c|cccccc|} 
\hline
Model & Module & CLIP & PSNR & SSIM & MSSSIM & FID & IS \\ 
\hline
\multirow{5}{*}{SD-v1-5} & Encoding & \bf 33.04 & 14.33 & 0.255 & 0.538 & \bf 174.8& 17.69\\
 & Unet & 33.44 & 13.53& \bf 0.219 & \bf 0.495 & 172.3& \bf 14.82\\
 & Decoding & 33.69 & \bf 13.29 & 0.248 & 0.520& 166.3& 21.65\\
 & Gaussian & 34.31 & 15.95 & 0.316 &0.628 & 161.1& 15.78\\
 & Benign &  34.40 & $\infty$ & 1.000 & 1.000 &156.4 &22.45 \\
\hline
\multirow{5}{*}{SD-v2-1} & Encoding & \bf 33.62 & 14.24 & 0.268 & 0.542 & \bf 184.3& \bf 13.74\\
 & Unet & 33.98 & \bf 13.75 & \bf 0.252 & \bf 0.524 & 174.1 & 17.83 \\
 & Decoding & 34.20 & 13.86 & 0.270 & 0.542 & 168.7& 19.00\\
 & Gaussian & 34.66 & 15.97 & 0.334 & 0.635& 155.7& 22.88\\
 & Benign & 34.88 & $\infty$ & 1.000 & 1.000  & 153.0& 23.56\\
\hline 
\end{tabular}}
\label{tablea3}
\end{table}

\begin{table}
\caption{The model-transfer attacking performance against other image inpainting diffusion models. The best results are marked in bold.}
\centering

\begin{tabular}{|c|c|cc|} 
\hline
Model & Module & SD-v1-5 & SD-v2-1 \\ 
\hline
\multirow{3}{*}{SD-v1-5} & Encoding & \bf 33.04 & \bf 33.61\\
 & Unet & 33.44 & 33.87\\
 & Decoding & 33.69 & 34.31\\

\hline
\multirow{3}{*}{SD-v2-1} & Encoding & \bf 33.61 & \bf 33.62\\
 & Unet & 33.72 & 33.98\\
 & Decoding & 33.87 & 34.20\\
 
\hline 
\end{tabular}
\label{tablea5}
\end{table}

Besides, we illustrate the model-transfer performance in Table \ref{tablea5}. We can see that the transfer attack can achieve better performance than the Gaussian noise in Table \ref{table3}. Therefore, the adversarial examples of image inpainting models still transfer across different models. Furthermore, the adversarial images crafted from SD-v1-5 and SD-v2-1 can achieve similar performance on attacking SD-v2-1, which means that the adversarial examples from SD-v1-5 are as effective as the ones from SD-v2-1 on misleading SD-v2-1. However, the adversarial examples crafted from SD-v2-1 cannot mislead SD-v1-5 well. Thus, the image inpainting latent diffusion models suffer from the same robustness problem as image variation models. The defects from SD-v1 are inherited by SD-v2, which also includes new defects.

\clearpage
\section{Qualitative Results}
\label{d}

We add more white-box qualitative results against the SD-v1-5 image variation model in Figure \ref{vis_2}. We also include the corresponding black-box qualitative results and the results against defense mechanisms in Figure \ref{vis_4}.

\begin{figure}[htbp]
\ \ \ \ \ \ \ \ \ \ Clean\ \ \ \ \ \ \ \ \ \ \ \ \ \ \ \ \ Generate \ \ \ \ \ \ \ \ \ \ \ \ \ \ \ Encoding\ \ \ \ \ \ \ \ \ \ \ \ \ \ \ \ \ Unet\ \ \ \ \ \ \ \ \ \ \ \ \ \ \ \ \ Decoding \\
\centering
\begin{minipage}[t]{0.18\textwidth}
\centering
\includegraphics[width=2.5cm]{variation/ori_image7.jpg}
\end{minipage}
\begin{minipage}[t]{0.18\textwidth}
\centering
\includegraphics[width=2.5cm]{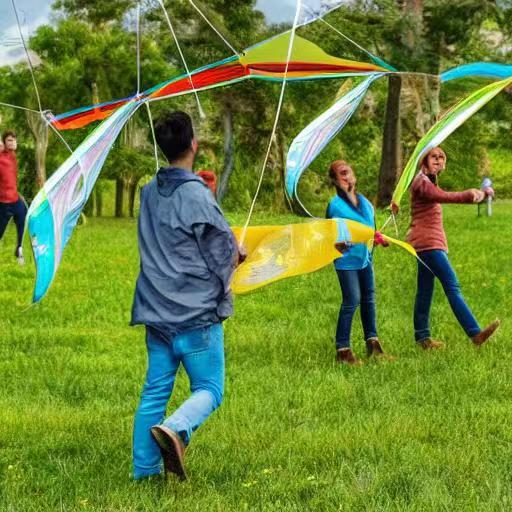}
\end{minipage}
\begin{minipage}[t]{0.18\textwidth}
\centering
\includegraphics[width=2.5cm]{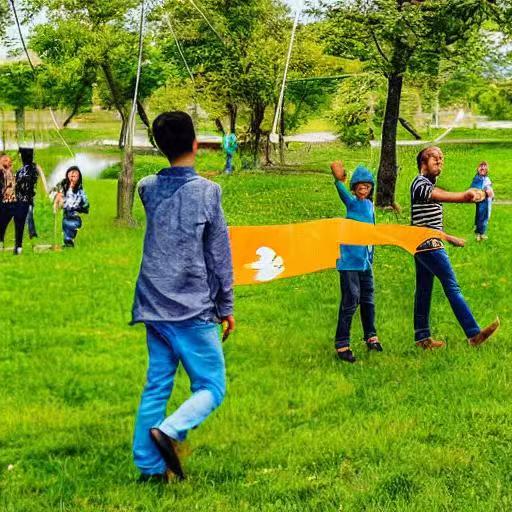}
\end{minipage}
\begin{minipage}[t]{0.18\textwidth}
\centering
\includegraphics[width=2.5cm]{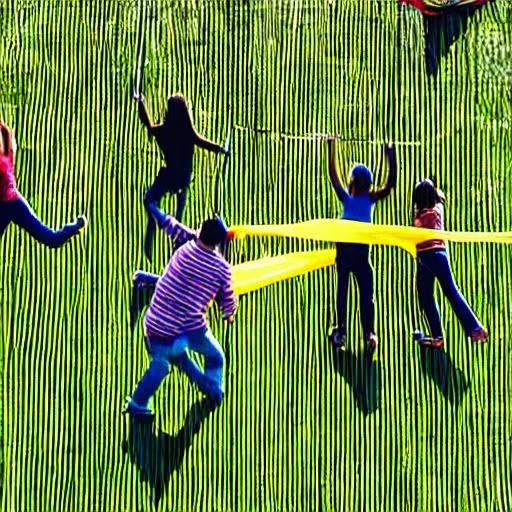}
\end{minipage}
\begin{minipage}[t]{0.18\textwidth}
\centering
\includegraphics[width=2.5cm]{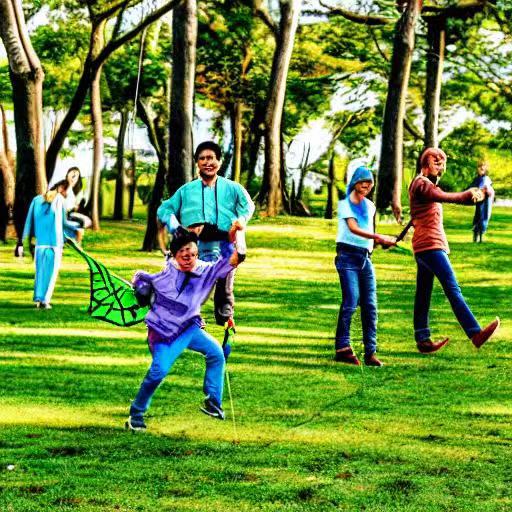}
\end{minipage}
\\"A group of people enjoying a bright kite flying in the recreational area." \\
\begin{minipage}[t]{0.18\textwidth}
\centering
\includegraphics[width=2.5cm]{variation/ori_image6.jpg}
\end{minipage}
\begin{minipage}[t]{0.18\textwidth}
\centering
\includegraphics[width=2.5cm]{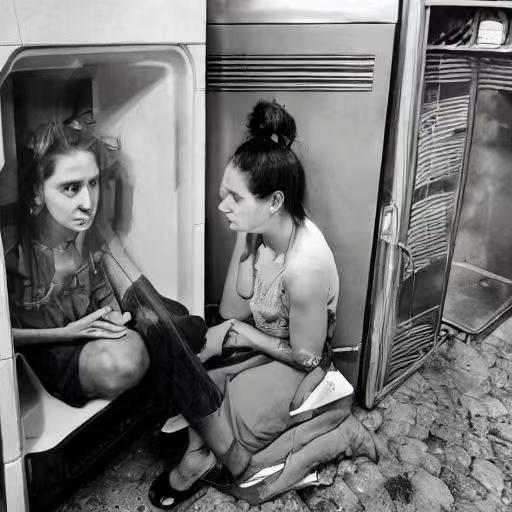}
\end{minipage}
\begin{minipage}[t]{0.18\textwidth}
\centering
\includegraphics[width=2.5cm]{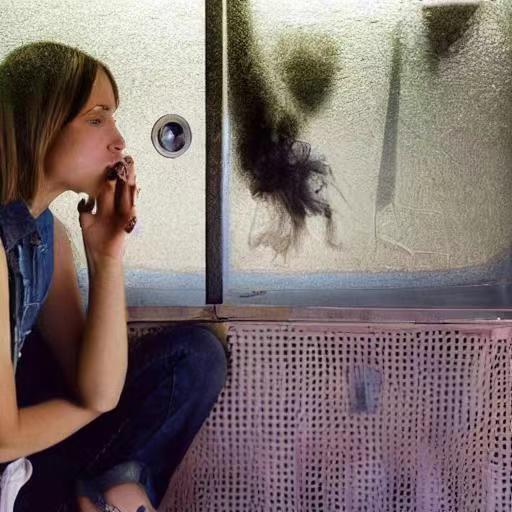}
\end{minipage}
\begin{minipage}[t]{0.18\textwidth}
\centering
\includegraphics[width=2.5cm]{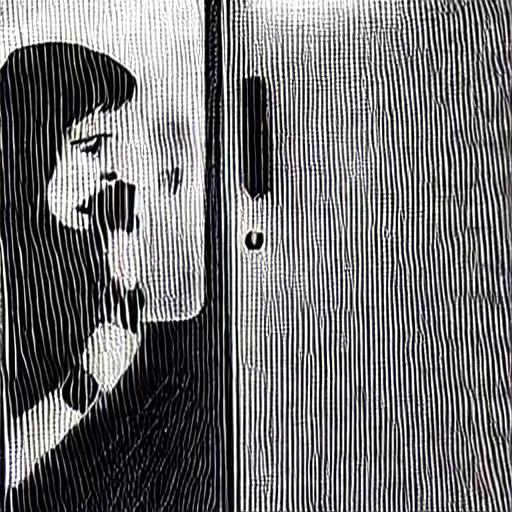}
\end{minipage}
\begin{minipage}[t]{0.18\textwidth}
\centering
\includegraphics[width=2.5cm]{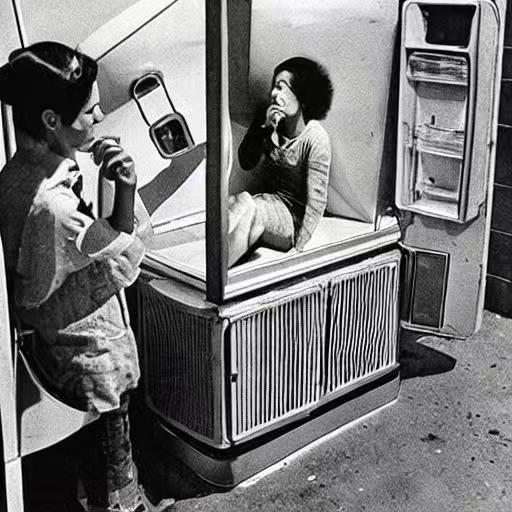}
\end{minipage}
\\"Near a girl sitting inside a refrigerator." \\
\begin{minipage}[t]{0.18\textwidth}
\centering
\includegraphics[width=2.5cm]{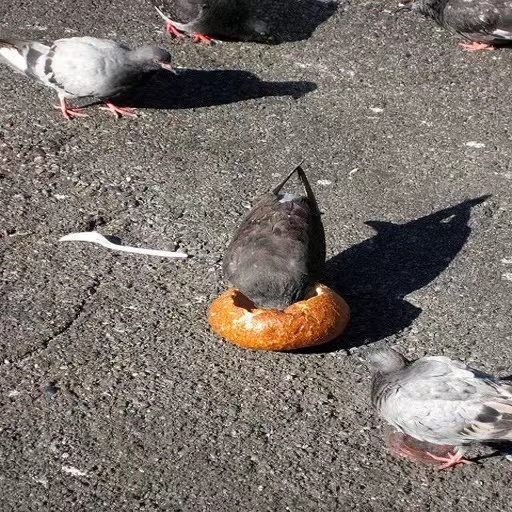}
\end{minipage}
\begin{minipage}[t]{0.18\textwidth}
\centering
\includegraphics[width=2.5cm]{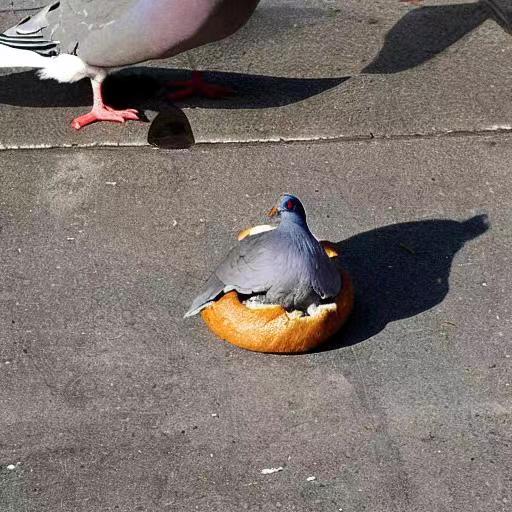}
\end{minipage}
\begin{minipage}[t]{0.18\textwidth}
\centering
\includegraphics[width=2.5cm]{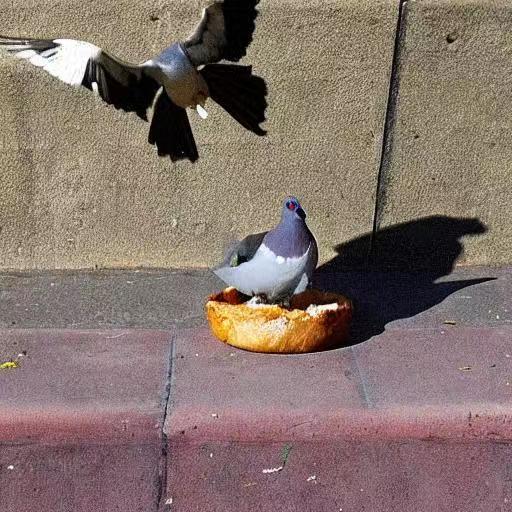}
\end{minipage}
\begin{minipage}[t]{0.18\textwidth}
\centering
\includegraphics[width=2.5cm]{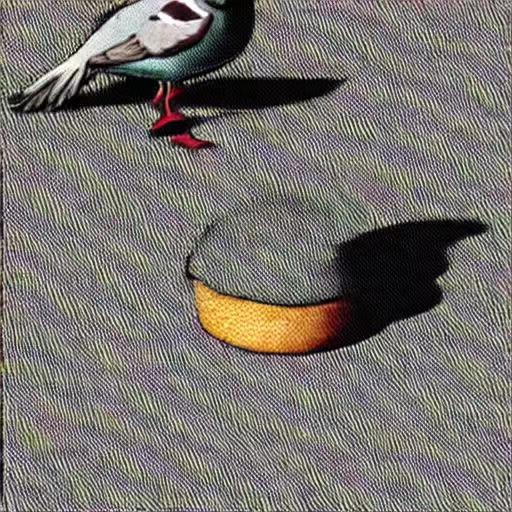}
\end{minipage}
\begin{minipage}[t]{0.18\textwidth}
\centering
\includegraphics[width=2.5cm]{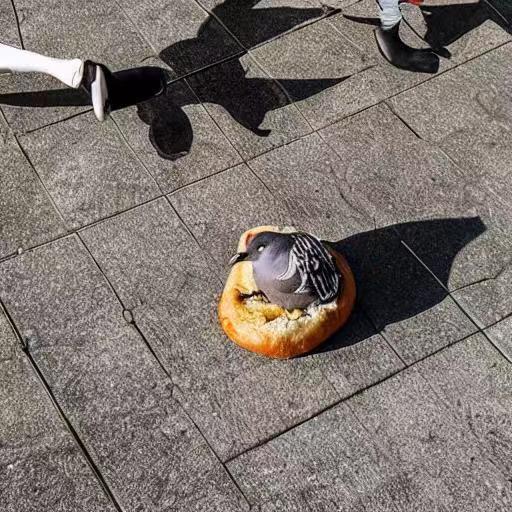}
\end{minipage}
\\"The pigeon is dining from a bread bowl, savoring every bite." \\
\begin{minipage}[t]{0.18\textwidth}
\centering
\includegraphics[width=2.5cm]{variation/ori_image3.jpg}
\end{minipage}
\begin{minipage}[t]{0.18\textwidth}
\centering
\includegraphics[width=2.5cm]{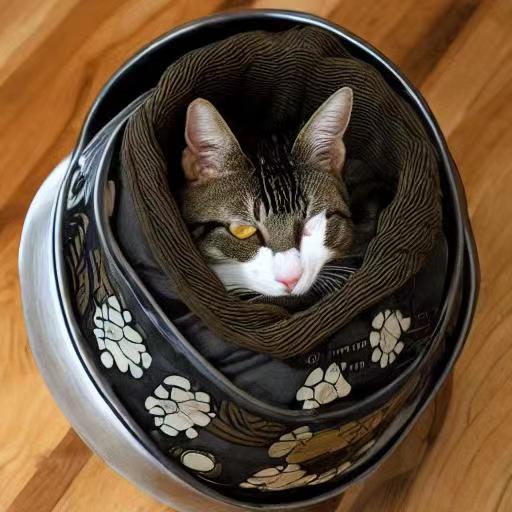}
\end{minipage}
\begin{minipage}[t]{0.18\textwidth}
\centering
\includegraphics[width=2.5cm]{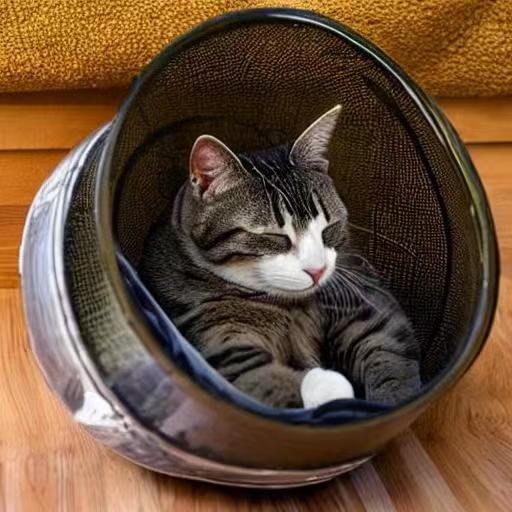}
\end{minipage}
\begin{minipage}[t]{0.18\textwidth}
\centering
\includegraphics[width=2.5cm]{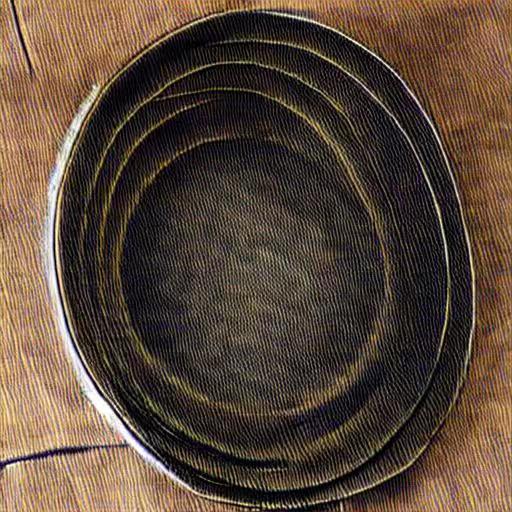}
\end{minipage}
\begin{minipage}[t]{0.18\textwidth}
\centering
\includegraphics[width=2.5cm]{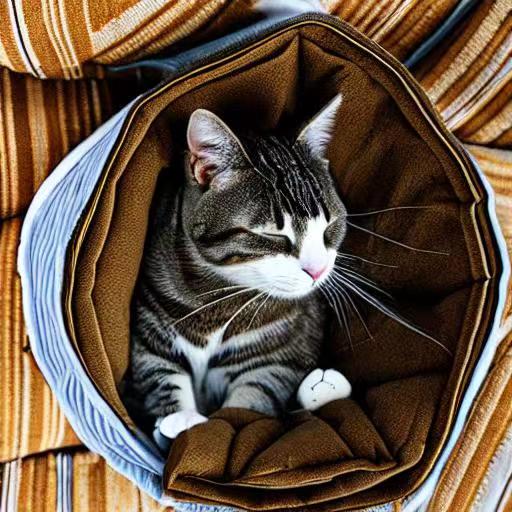}
\end{minipage}
\\"A bowl on the ground is transformed into a cozy bed for a cat catching some z's." \\
\begin{minipage}[t]{0.18\textwidth}
\centering
\includegraphics[width=2.5cm]{variation/ori_image2.jpg}
\end{minipage}
\begin{minipage}[t]{0.18\textwidth}
\centering
\includegraphics[width=2.5cm]{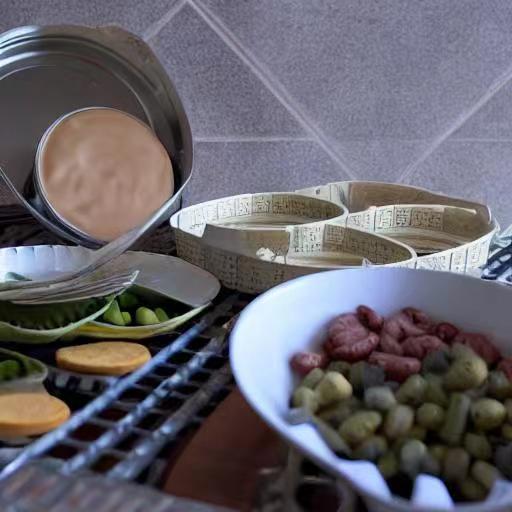}
\end{minipage}
\begin{minipage}[t]{0.18\textwidth}
\centering
\includegraphics[width=2.5cm]{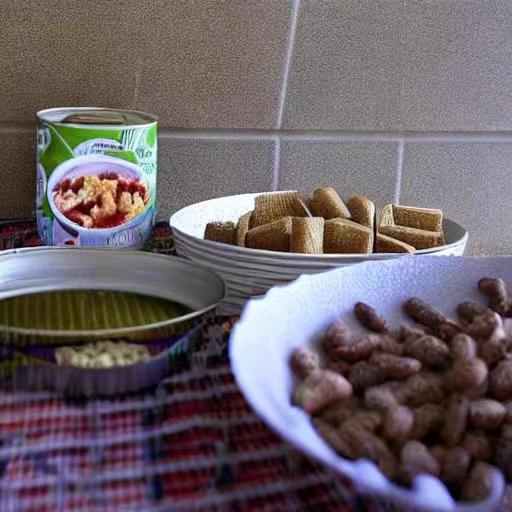}
\end{minipage}
\begin{minipage}[t]{0.18\textwidth}
\centering
\includegraphics[width=2.5cm]{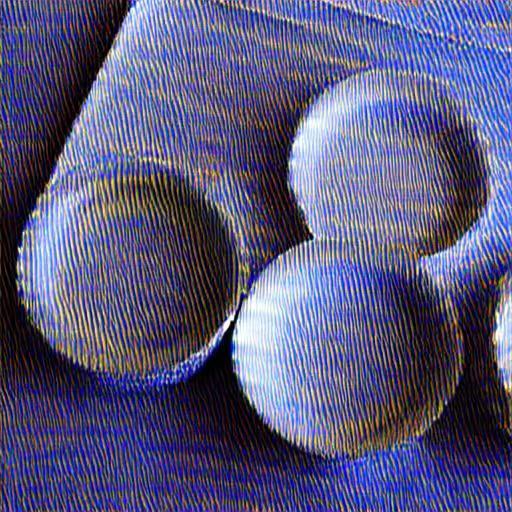}
\end{minipage}
\begin{minipage}[t]{0.18\textwidth}
\centering
\includegraphics[width=2.5cm]{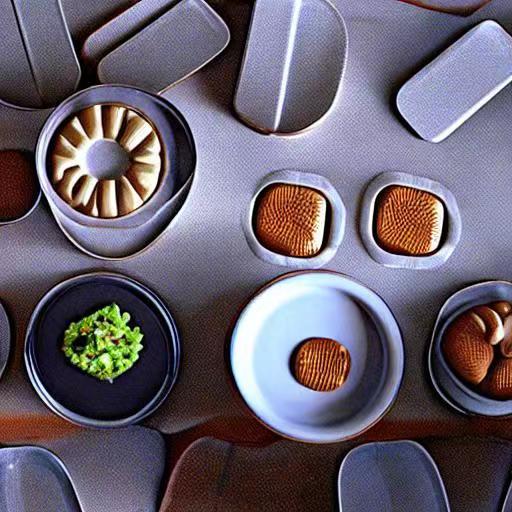}
\end{minipage}
\\ "Trays with wraps next to a can of potted meat." \\
\begin{minipage}[t]{0.18\textwidth}
\centering
\includegraphics[width=2.5cm]{variation/ori_image5.jpg}
\end{minipage}
\begin{minipage}[t]{0.18\textwidth}
\centering
\includegraphics[width=2.5cm]{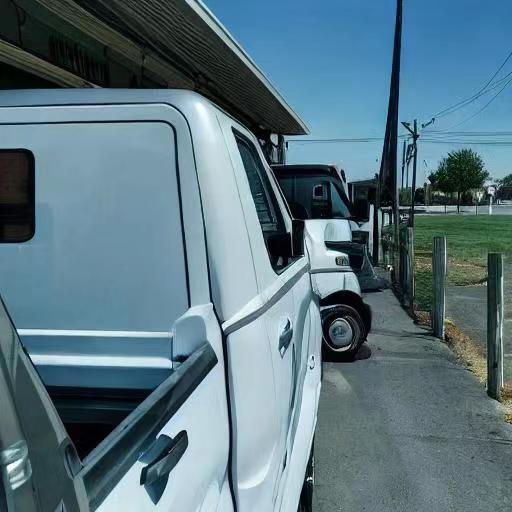}
\end{minipage}
\begin{minipage}[t]{0.18\textwidth}
\centering
\includegraphics[width=2.5cm]{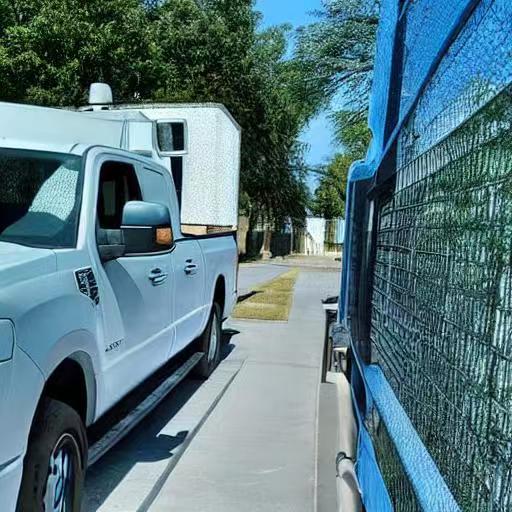}
\end{minipage}
\begin{minipage}[t]{0.18\textwidth}
\centering
\includegraphics[width=2.5cm]{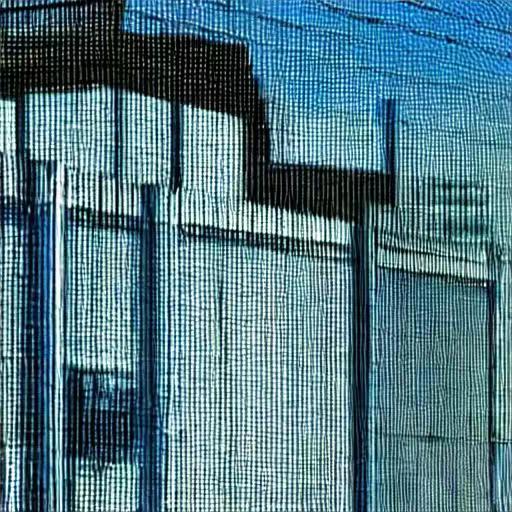}
\end{minipage}
\begin{minipage}[t]{0.18\textwidth}
\centering
\includegraphics[width=2.5cm]{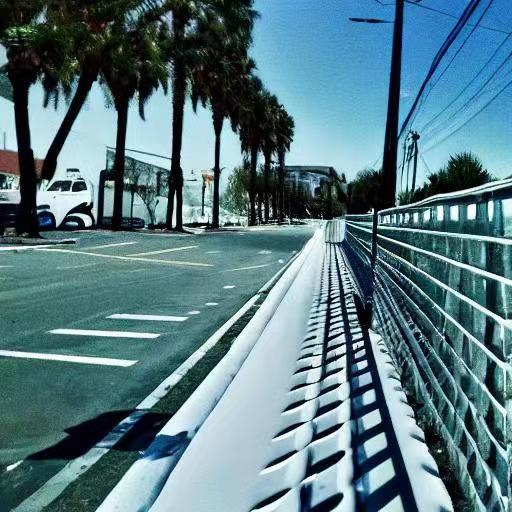}
\end{minipage}
\\"A white truck parked next to a sidewalk near a fence." \\
%\caption{More qualitative results of adversarial images and their output of the SD-v1-5 image variation model.}
\end{figure}
\clearpage

\begin{figure}[htbp]
%\ \ \ \ \ \ \ \ \ \ Clean\ \ \ \ \ \ \ \ \ \ \ \ \ \ \ \ \ Generate \ \ \ \ \ \ \ \ \ \ \ \ \ \ \ Encoding\ \ \ \ \ \ \ \ \ \ \ \ \ \ \ \ \ Unet\ \ \ \ \ \ \ \ \ \ \ \ \ \ \ \ \ Decoding \\
\centering
\begin{minipage}[t]{0.18\textwidth}
\centering
\includegraphics[width=2.5cm]{variation/ori_image8.jpg}
\end{minipage}
\begin{minipage}[t]{0.18\textwidth}
\centering
\includegraphics[width=2.5cm]{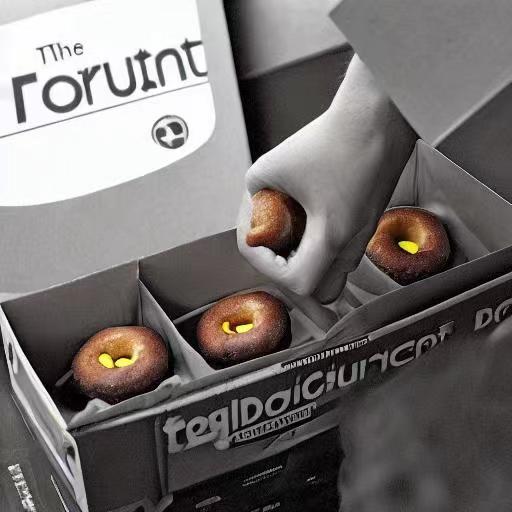}
\end{minipage}
\begin{minipage}[t]{0.18\textwidth}
\centering
\includegraphics[width=2.5cm]{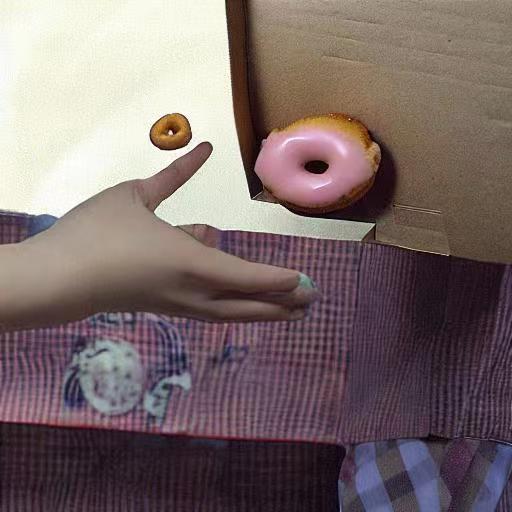}
\end{minipage}
\begin{minipage}[t]{0.18\textwidth}
\centering
\includegraphics[width=2.5cm]{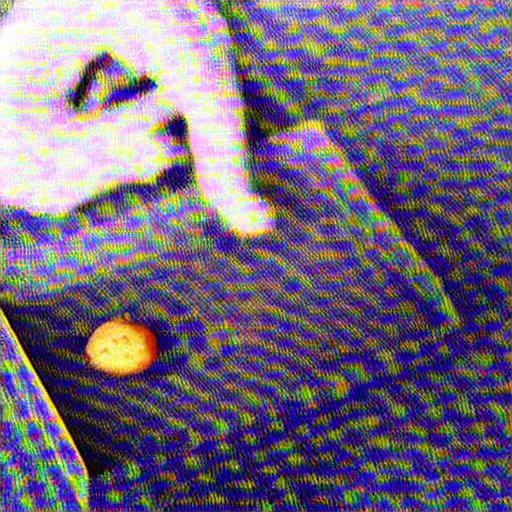}
\end{minipage}
\begin{minipage}[t]{0.18\textwidth}
\centering
\includegraphics[width=2.5cm]{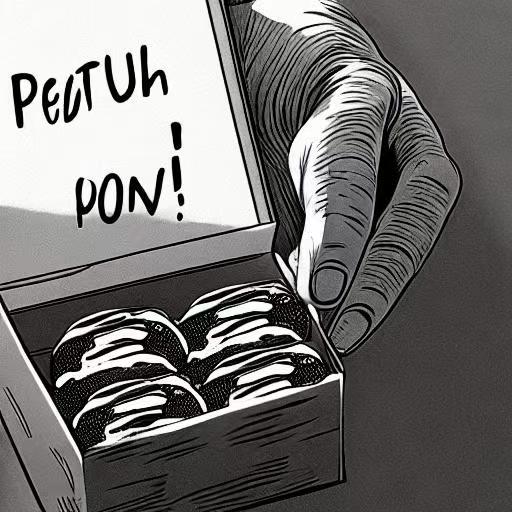}
\end{minipage}
\\ "In the image, a person is reaching for a doughnut from an open box." \\
\begin{minipage}[t]{0.18\textwidth}
\centering
\includegraphics[width=2.5cm]{variation/ori_image10.jpg}
\end{minipage}
\begin{minipage}[t]{0.18\textwidth}
\centering
\includegraphics[width=2.5cm]{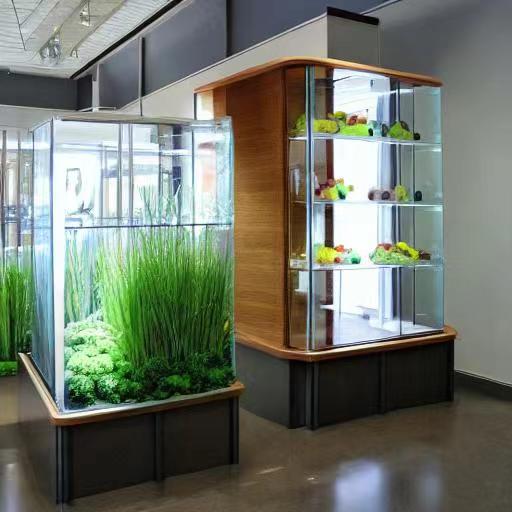}
\end{minipage}
\begin{minipage}[t]{0.18\textwidth}
\centering
\includegraphics[width=2.5cm]{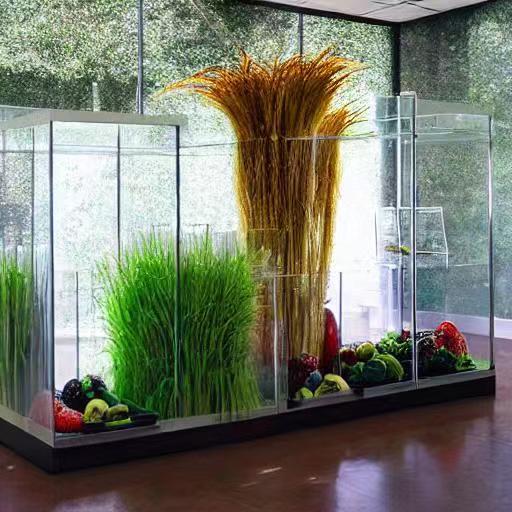}
\end{minipage}
\begin{minipage}[t]{0.18\textwidth}
\centering
\includegraphics[width=2.5cm]{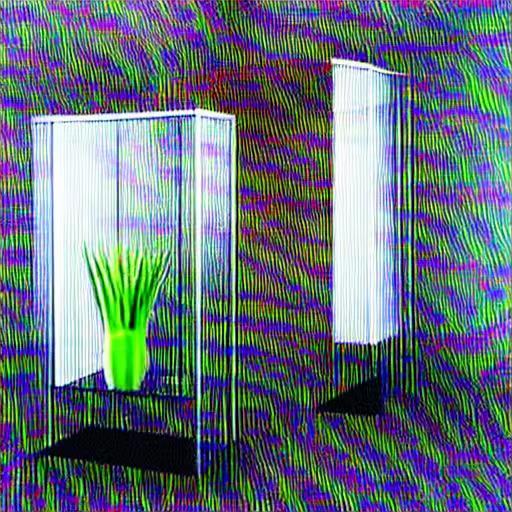}
\end{minipage}
\begin{minipage}[t]{0.18\textwidth}
\centering
\includegraphics[width=2.5cm]{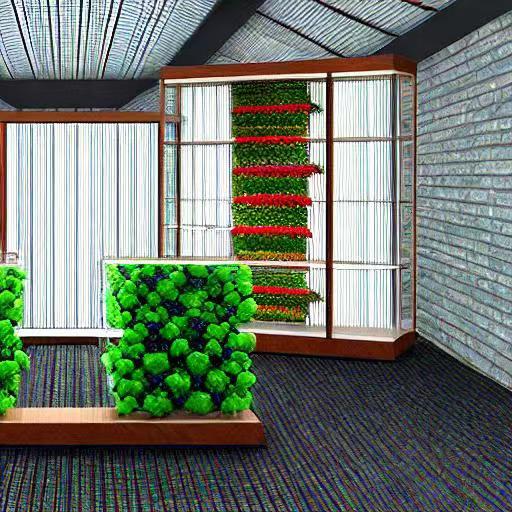}
\end{minipage}
\\"A display case with tall glass walls featuring a mix of fruit and grass." \\
\begin{minipage}[t]{0.18\textwidth}
\centering
\includegraphics[width=2.5cm]{variation/ori_image9.jpg}
\end{minipage}
\begin{minipage}[t]{0.18\textwidth}
\centering
\includegraphics[width=2.5cm]{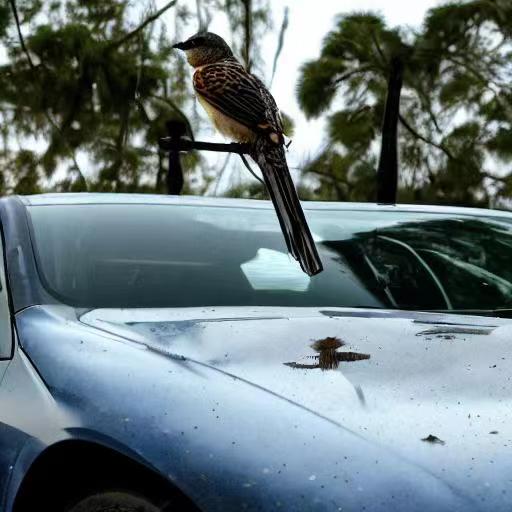}
\end{minipage}
\begin{minipage}[t]{0.18\textwidth}
\centering
\includegraphics[width=2.5cm]{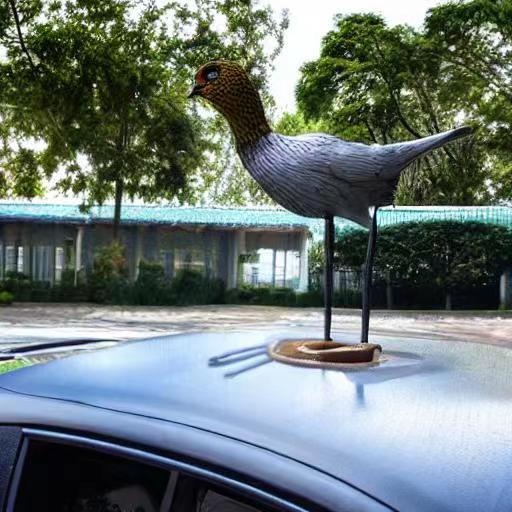}
\end{minipage}
\begin{minipage}[t]{0.18\textwidth}
\centering
\includegraphics[width=2.5cm]{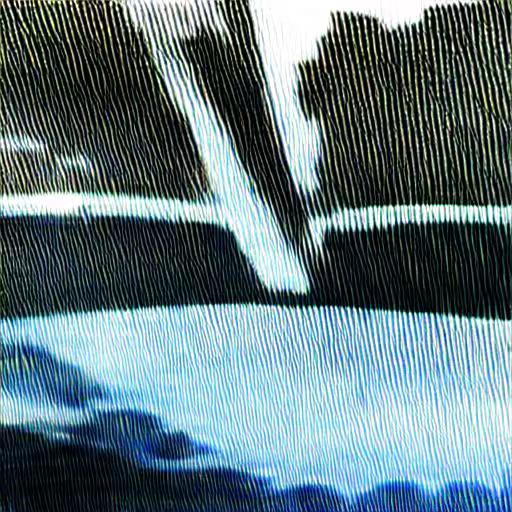}
\end{minipage}
\begin{minipage}[t]{0.18\textwidth}
\centering
\includegraphics[width=2.5cm]{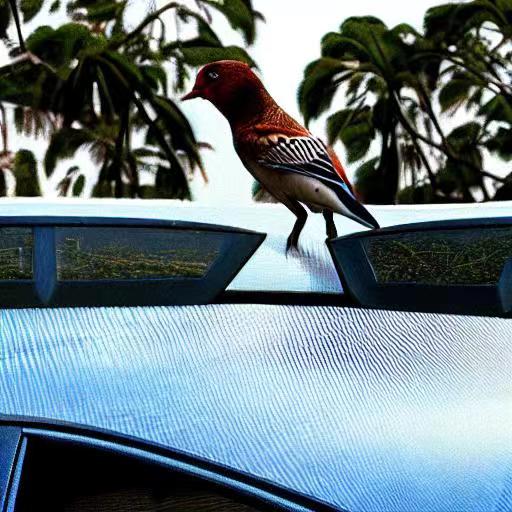}
\end{minipage}
\\" A car roof with a bird standing on it near a covered pavilion." \\
\caption{More qualitative results of adversarial images and their output of the SD-v1-5 image variation model.}
\label{vis_2}
\end{figure}

\begin{figure}[htbp]
\ \ \ \ \ \ \ SD-v1-4\ \ \ \ \ \ \ \ \ \ \ \ \ \ \ \ \ \ \ \ \ \ \ \ SD-v2-1 \ \ \ \ \ \ \ \ \ \ \ \ \ \ \ \ \ \ \ \ \ \ \ \ R \& P\ \ \ \ \ \ \ \ \ \ \ \ \ \ \ \ \ \ \ \ \ \ \ \ \ \ \ \ JPEG \\
\centering
\begin{minipage}[t]{0.23\textwidth}
\centering
\includegraphics[width=2.5cm]{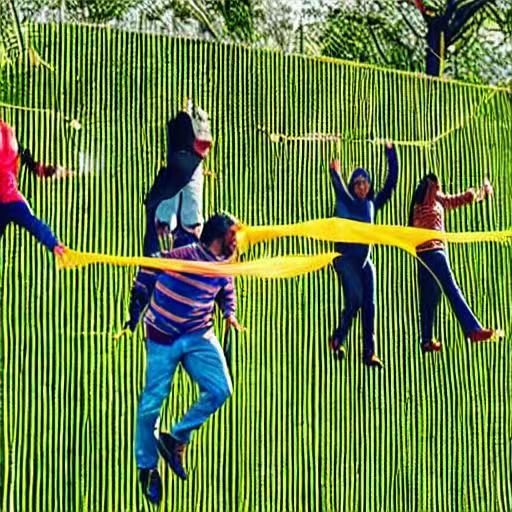}
\end{minipage}
\begin{minipage}[t]{0.23\textwidth}
\centering
\includegraphics[width=2.5cm]{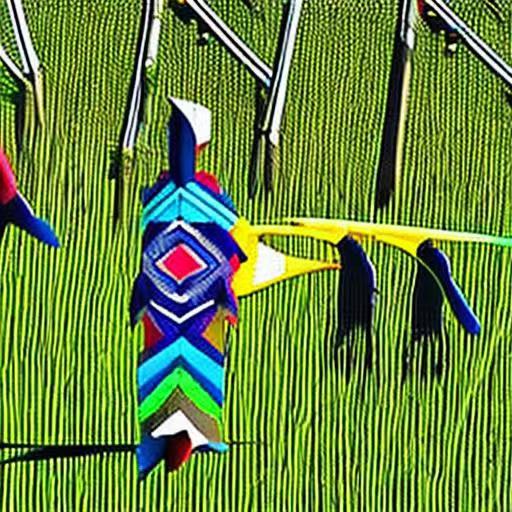}
\end{minipage}
\begin{minipage}[t]{0.23\textwidth}
\centering
\includegraphics[width=2.5cm]{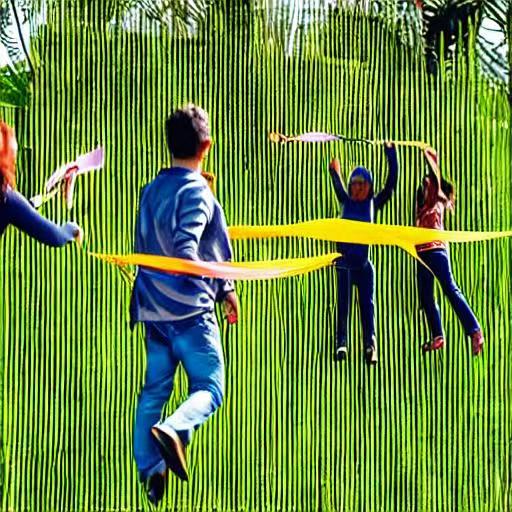}
\end{minipage}
\begin{minipage}[t]{0.23\textwidth}
\centering
\includegraphics[width=2.5cm]{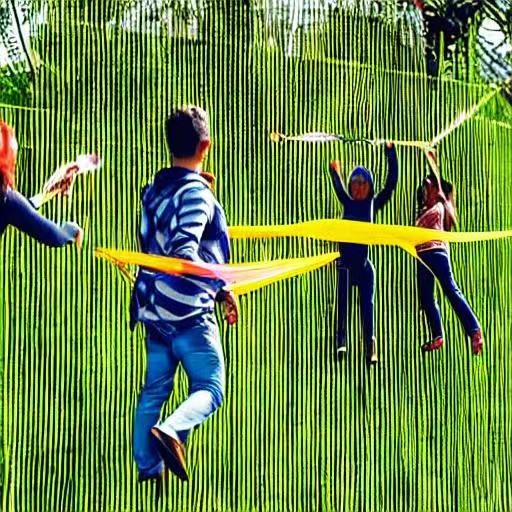}
\end{minipage}
\\ "A group of people enjoying a bright kite flying in the recreational area." \\
\begin{minipage}[t]{0.23\textwidth}
\centering
\includegraphics[width=2.5cm]{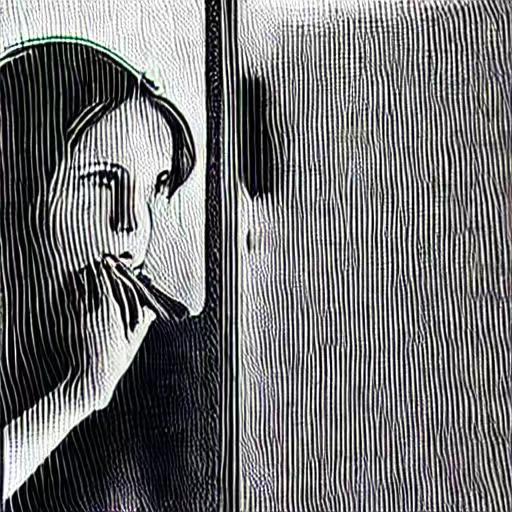}
\end{minipage}
\begin{minipage}[t]{0.23\textwidth}
\centering
\includegraphics[width=2.5cm]{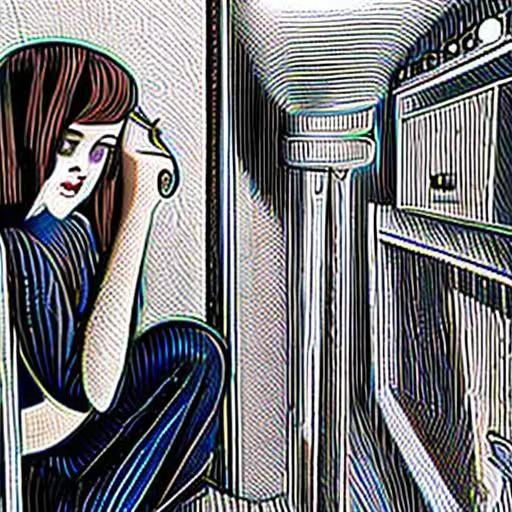}
\end{minipage}
\begin{minipage}[t]{0.23\textwidth}
\centering
\includegraphics[width=2.5cm]{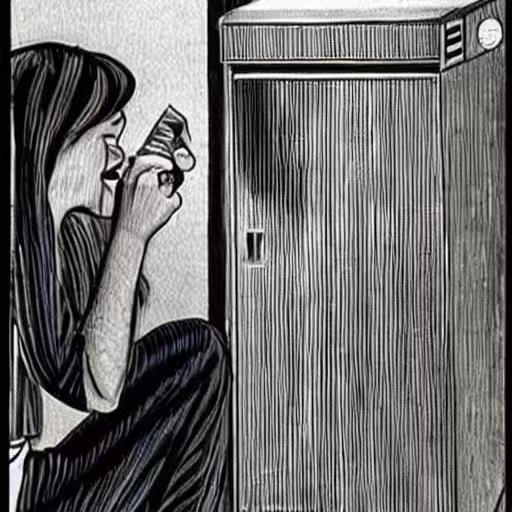}
\end{minipage}
\begin{minipage}[t]{0.23\textwidth}
\centering
\includegraphics[width=2.5cm]{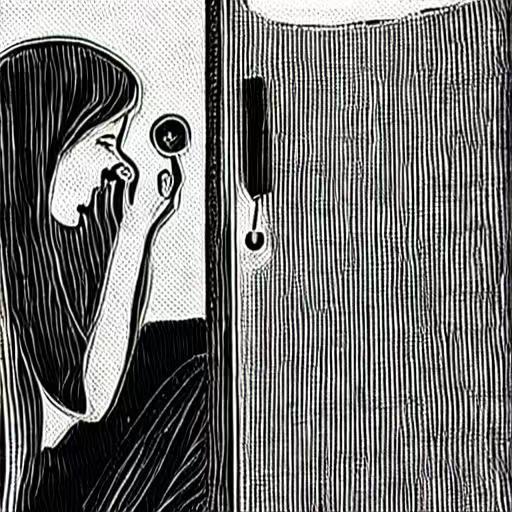}
\end{minipage}
\\ "Near a girl sitting inside a refrigerator." \\
\begin{minipage}[t]{0.23\textwidth}
\centering
\includegraphics[width=2.5cm]{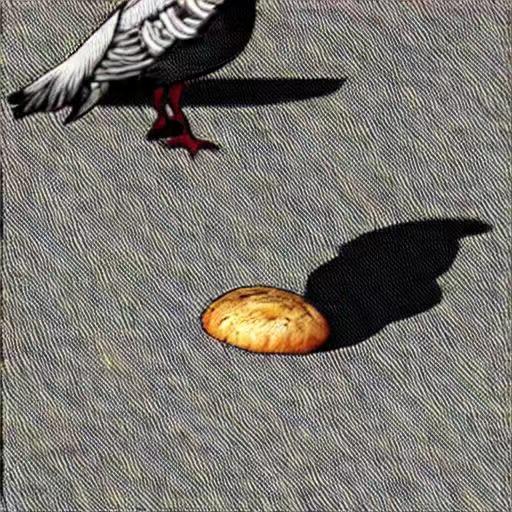}
\end{minipage}
\begin{minipage}[t]{0.23\textwidth}
\centering
\includegraphics[width=2.5cm]{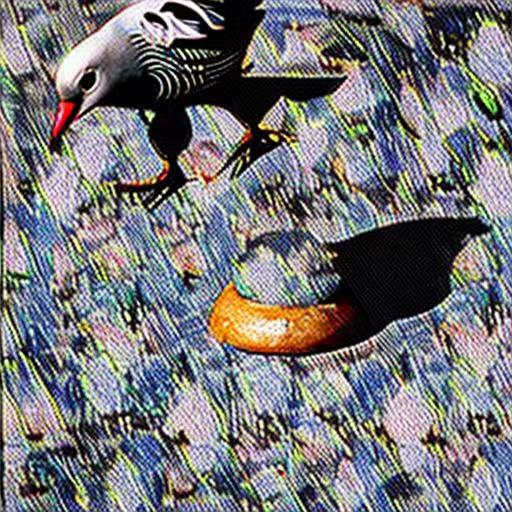}
\end{minipage}
\begin{minipage}[t]{0.23\textwidth}
\centering
\includegraphics[width=2.5cm]{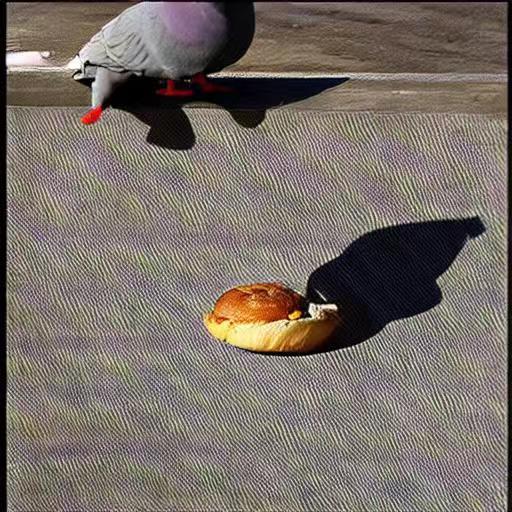}
\end{minipage}
\begin{minipage}[t]{0.23\textwidth}
\centering
\includegraphics[width=2.5cm]{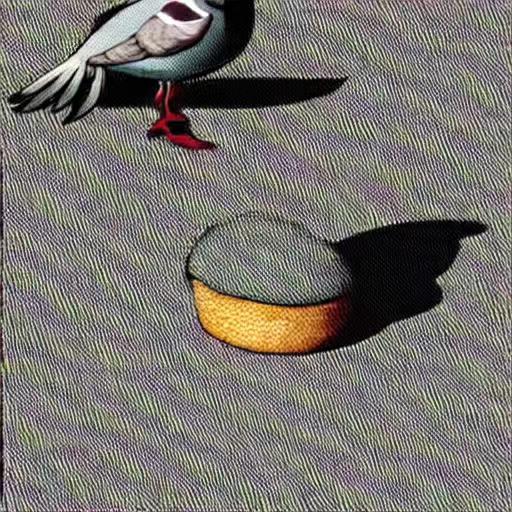}
\end{minipage}
\\ "The pigeon is dining from a bread bowl, savoring every bite." \\
%\caption{More qualitative results of generated images by model-transfer attacks on SD-v1-4 and SD-v2-1, and the generated adversarial images of the SD-v1-5 against defense mechanisms.}
\end{figure}
\clearpage   

\begin{figure}[t]
%\ \ \ \ \ \ \ SD-v1-4\ \ \ \ \ \ \ \ \ \ \ \ \ \ \ \ \ \ \ \ \ \ \ \ SD-v2-1 \ \ \ \ \ \ \ \ \ \ \ \ \ \ \ \ \ \ \ \ \ \ \ \ R \& P\ \ \ \ \ \ \ \ \ \ \ \ \ \ \ \ \ \ \ \ \ \ \ \ \ \ \ \ JPEG \\
\centering
\begin{minipage}[t]{0.23\textwidth}
\centering
\includegraphics[width=2.5cm]{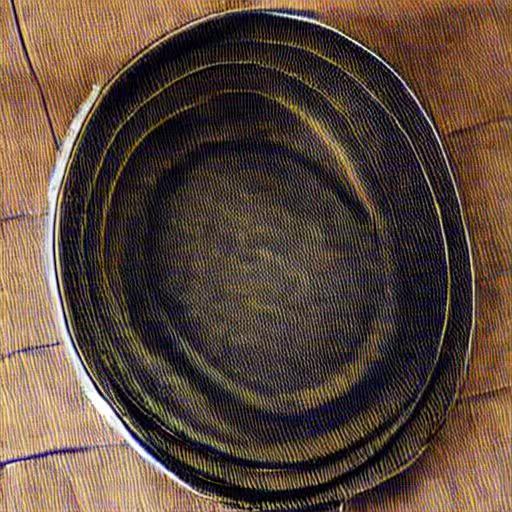}
\end{minipage}
\begin{minipage}[t]{0.23\textwidth}
\centering
\includegraphics[width=2.5cm]{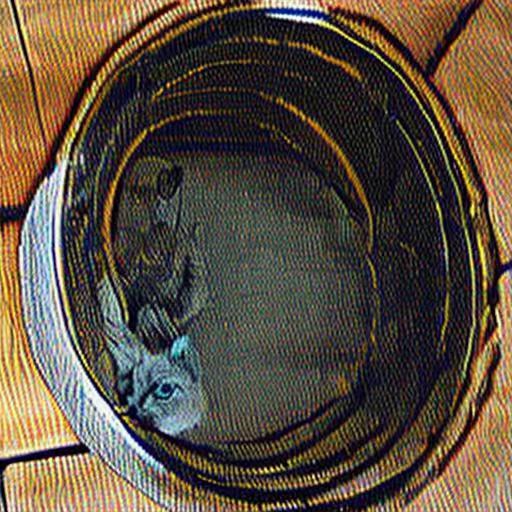}
\end{minipage}
\begin{minipage}[t]{0.23\textwidth}
\centering
\includegraphics[width=2.5cm]{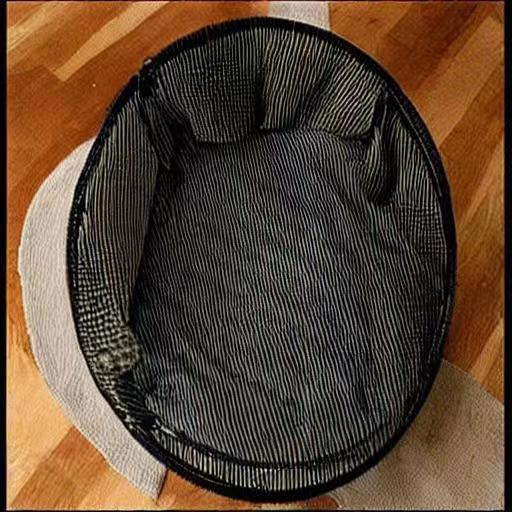}
\end{minipage}
\begin{minipage}[t]{0.23\textwidth}
\centering
\includegraphics[width=2.5cm]{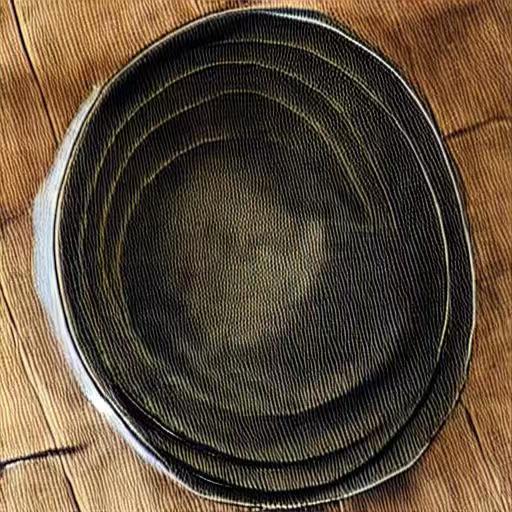}
\end{minipage}
\\ "A bowl on the ground is transformed into a cozy bed for a cat catching some z's." \\
\begin{minipage}[t]{0.23\textwidth}
\centering
\includegraphics[width=2.5cm]{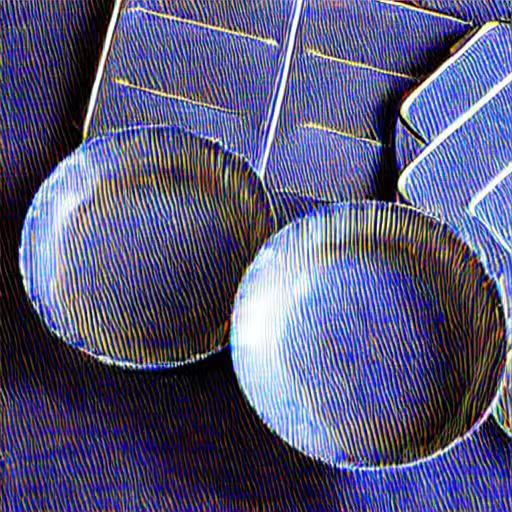}
\end{minipage}
\begin{minipage}[t]{0.23\textwidth}
\centering
\includegraphics[width=2.5cm]{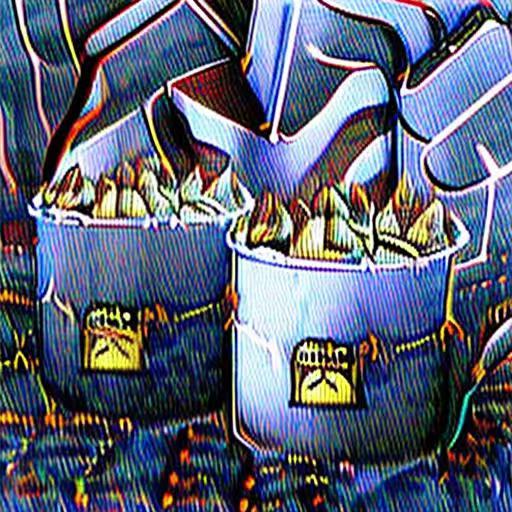}
\end{minipage}
\begin{minipage}[t]{0.23\textwidth}
\centering
\includegraphics[width=2.5cm]{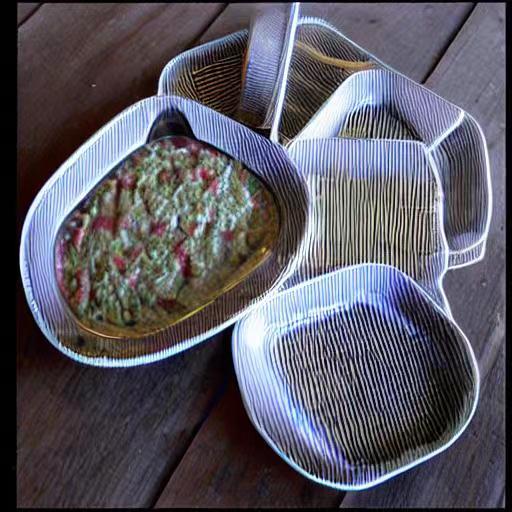}
\end{minipage}
\begin{minipage}[t]{0.23\textwidth}
\centering
\includegraphics[width=2.5cm]{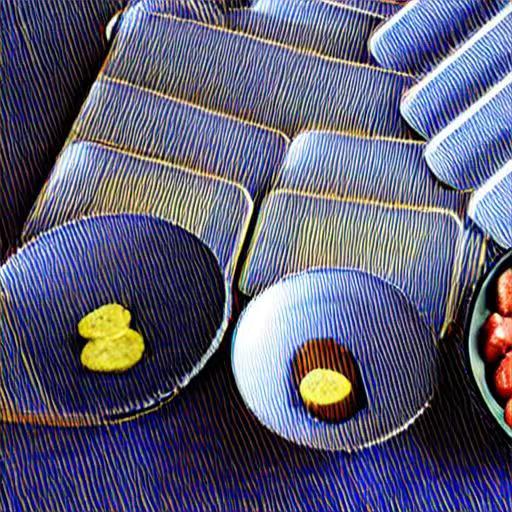}
\end{minipage}
\\ "Trays with wraps next to a can of potted meat." \\
\begin{minipage}[t]{0.23\textwidth}
\centering
\includegraphics[width=2.5cm]{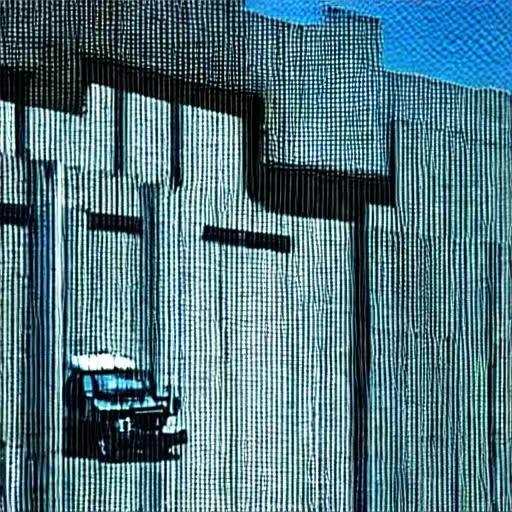}
\end{minipage}
\begin{minipage}[t]{0.23\textwidth}
\centering
\includegraphics[width=2.5cm]{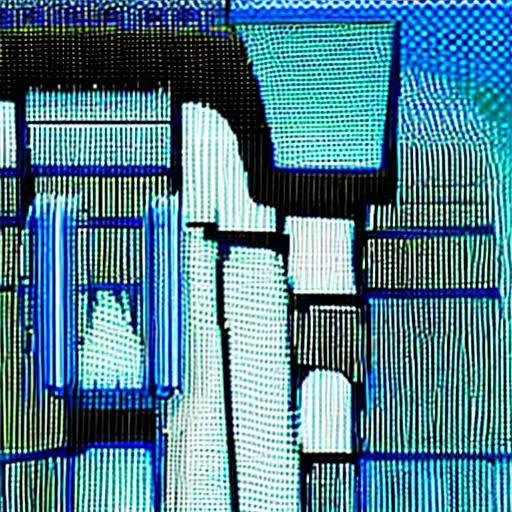}
\end{minipage}
\begin{minipage}[t]{0.23\textwidth}
\centering
\includegraphics[width=2.5cm]{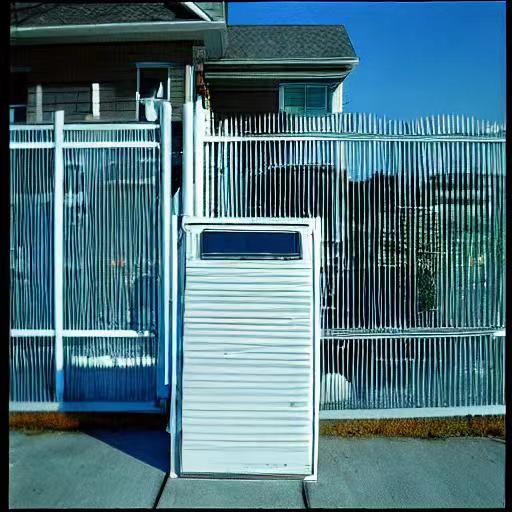}
\end{minipage}
\begin{minipage}[t]{0.23\textwidth}
\centering
\includegraphics[width=2.5cm]{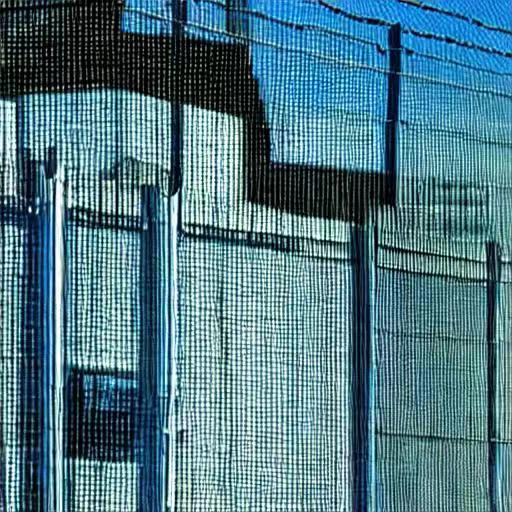}
\end{minipage}
\\ " A white truck parked next to a sidewalk near a fence." \\
\begin{minipage}[t]{0.23\textwidth}
\centering
\includegraphics[width=2.5cm]{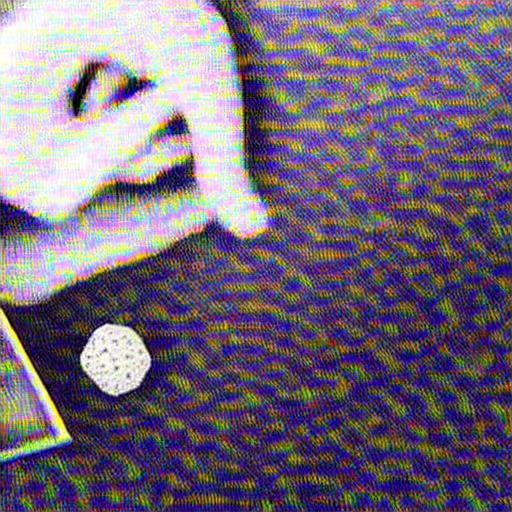}
\end{minipage}
\begin{minipage}[t]{0.23\textwidth}
\centering
\includegraphics[width=2.5cm]{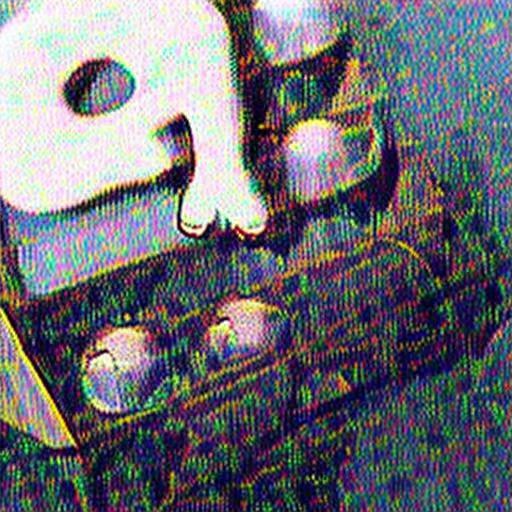}
\end{minipage}
\begin{minipage}[t]{0.23\textwidth}
\centering
\includegraphics[width=2.5cm]{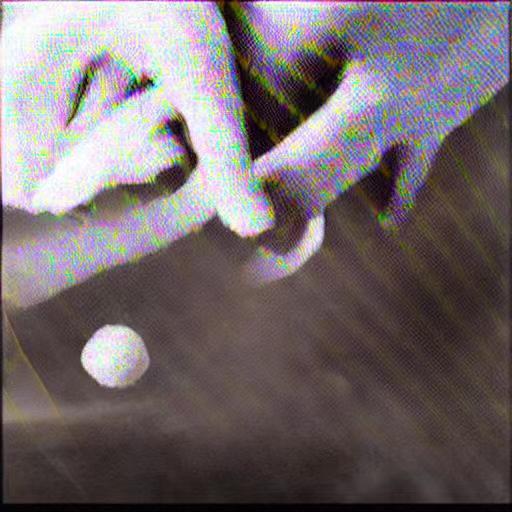}
\end{minipage}
\begin{minipage}[t]{0.23\textwidth}
\centering
\includegraphics[width=2.5cm]{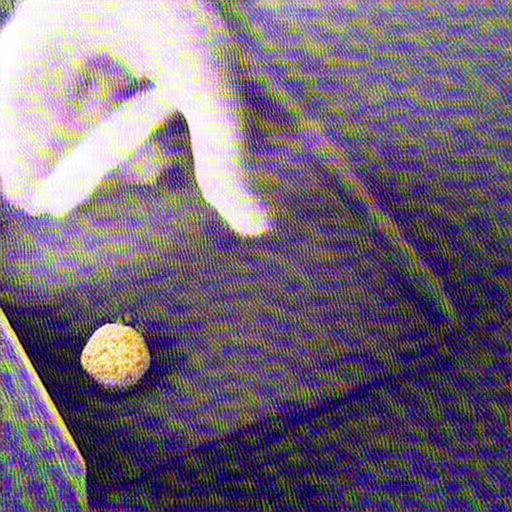}
\end{minipage}
\\ "In the image, a person is reaching for a doughnut from an open box." \\
\begin{minipage}[t]{0.23\textwidth}
\centering
\includegraphics[width=2.5cm]{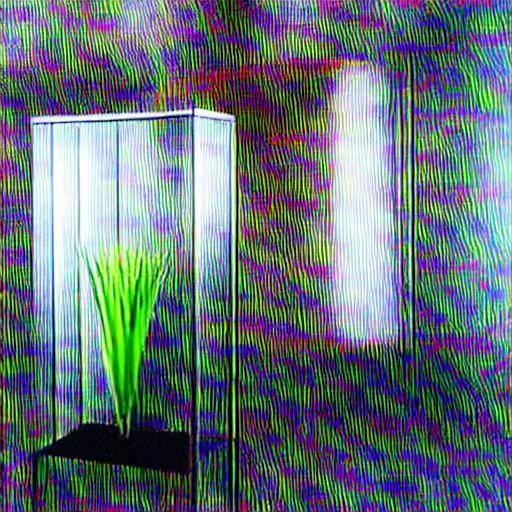}
\end{minipage}
\begin{minipage}[t]{0.23\textwidth}
\centering
\includegraphics[width=2.5cm]{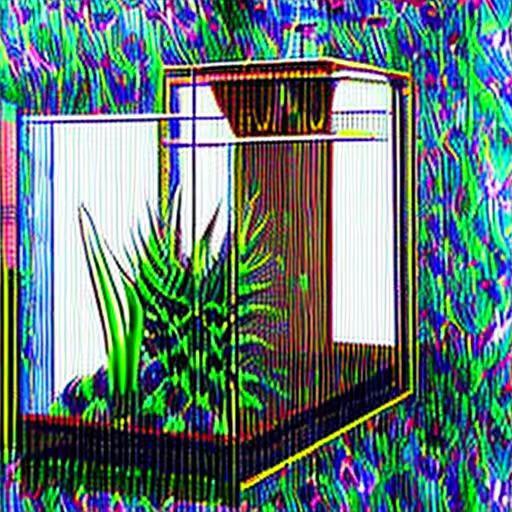}
\end{minipage}
\begin{minipage}[t]{0.23\textwidth}
\centering
\includegraphics[width=2.5cm]{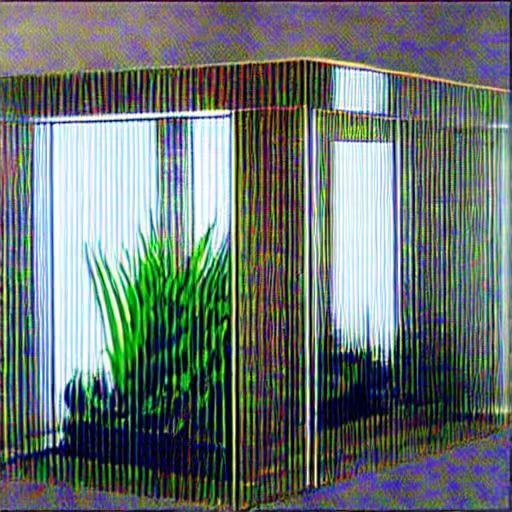}
\end{minipage}
\begin{minipage}[t]{0.23\textwidth}
\centering
\includegraphics[width=2.5cm]{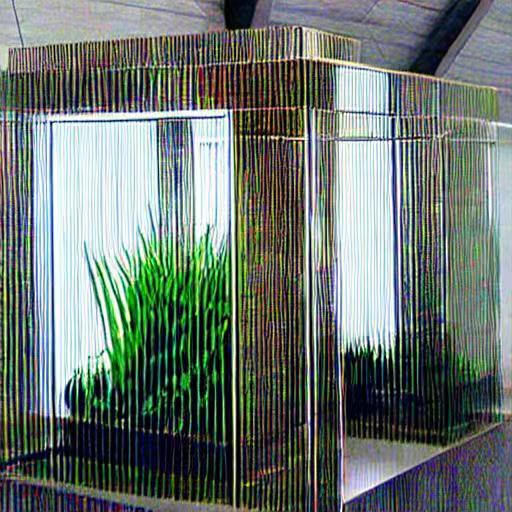}
\end{minipage}
\\ " A display case with tall glass walls featuring a mix of fruit and grass." \\
\begin{minipage}[t]{0.23\textwidth}
\centering
\includegraphics[width=2.5cm]{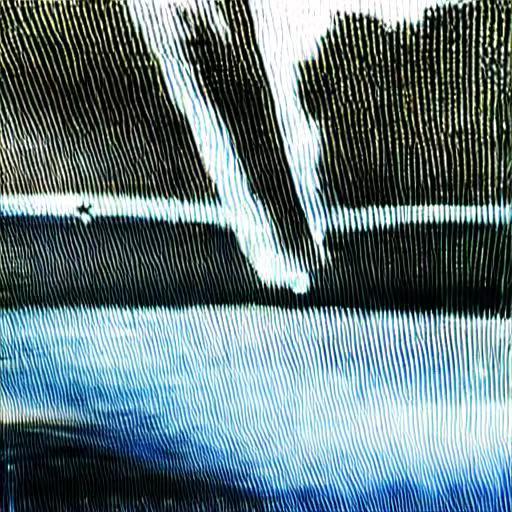}
\end{minipage}
\begin{minipage}[t]{0.23\textwidth}
\centering
\includegraphics[width=2.5cm]{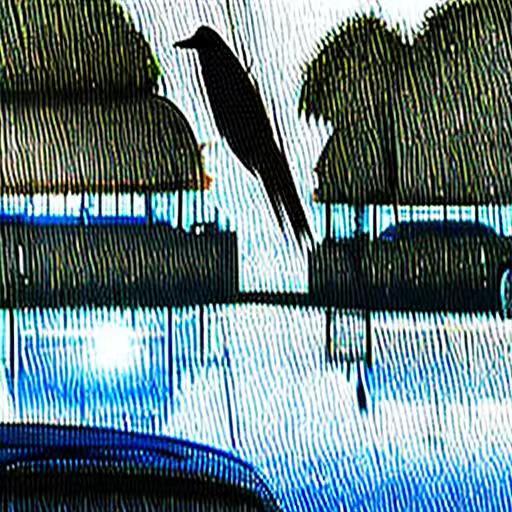}
\end{minipage}
\begin{minipage}[t]{0.23\textwidth}
\centering
\includegraphics[width=2.5cm]{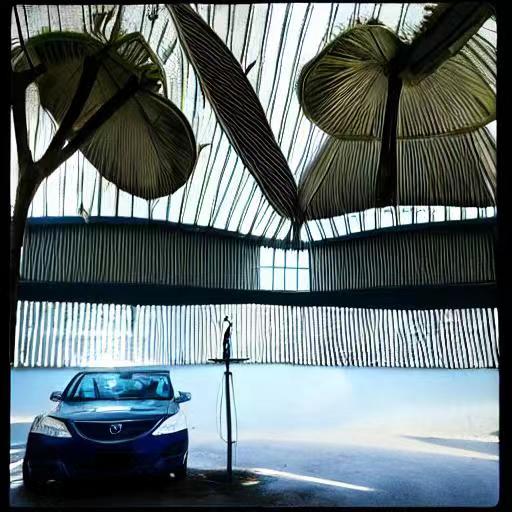}
\end{minipage}
\begin{minipage}[t]{0.23\textwidth}
\centering
\includegraphics[width=2.5cm]{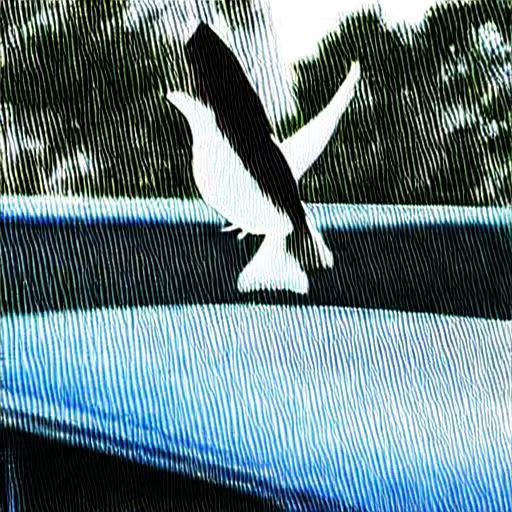}
\end{minipage}
\\ "A car roof with a bird standing on it near a covered pavilion." \\
\caption{More qualitative results of generated images by model-transfer attacks on SD-v1-4 and SD-v2-1, and generated adversarial images of the SD-v1-5 against defense mechanisms.}
\label{vis_4}
\end{figure}

\section{Broader Impacts \& Limitations}
\label{e}

Adversarial attacks can trigger bugs inside deep neural networks, and such technologies can be misused in the real world. However, it is important to understand the inner bugs inside the diffusion models. Our study can attract the attention of researchers and motivate the research on the robustness of diffusion models, like improving the robustness of the diffusion models and designing powerful defense mechanisms to avoid adversarial attacks. Moreover, adversarial attack strategies can be utilized in a good way to avoid intended malicious editing by the generative models for a harmonious society. 

The limitation of our work is that we propose to destroy all the internal features of the target module in the denoising process. However, we have no guarantee that attacking all the features is the optimal strategy. It is possible that better results are achievable by attacking part of the steps, but this would require tuning the step combination. We leave it to future work.

\end{document}